\documentclass[a4paper,12pt]{article}
\usepackage{amsfonts}
\usepackage[english]{babel}
\usepackage{graphicx}
\usepackage[usenames]{color}
\usepackage{colordvi}
\usepackage{amssymb}
\usepackage{amsmath}
\usepackage[round]{natbib}
\usepackage{multirow}
\usepackage{multicol}
\usepackage{ifthen}
\usepackage{ulem}
\usepackage{xcolor,eso-pic}
\usepackage{colordef}
\usepackage{tikz}
\usepackage{weiwER}  %% enhanced ER diagram drawing support based on pgf's ER library
{\quad ${\Box}$ \newline}
\def\proofname{Proof}

\def\remarkname{REMARK}
\newtheorem{theorem}{THEOREM}[section]

\newtheorem{assumption}{ASSUMPTION}[section]

\newtheorem{lemma}{LEMMA}[section]

\begin{document}
% Defined by Song, be CAREFUL when using it. Mostly specific paper related
\newcommand{\xy}{\{(X_i, Y_i)\}_{i=1}^n}
\newcommand{\lnh}{l_{n,h}}
\newcommand{\lng}{l_{n,g}}
\def\t{^{\scriptstyle{\mathsf{T}}} \hspace{-0.15em}}
\def\T{\top}
\def\ct{\mathcal{T}}
\def\cx{\mathcal{X}}
\def\ca{\mathcal{A}}
\def\ztt{Z_t^\T}
\def\tztt{\widetilde Z_t^\T}
\def\sfa{\scalefont{0.6}}
\def\jt1{T^{-1}}
\def\tT{1 \leqslant t \leqslant T}
\def\ta{\widetilde A}
\def\sfa{\scalefont{0.6}}
\def\sfb{\scalefont{1.6666666666}}
\def\del{\delta}
\def\al{\alpha}
\def\ti{\tilde}
\def\bb#1{\mathbb{#1}}
     \def\bbs{\ensuremath{{\bb S}}}
\def\t{^{\scriptstyle{\mathsf{T}}} \hspace{-0.15em}}

% from Ritov
\newcommand{\eqsplit}[2][*]%
  {\ifthenelse{\equal{#1}{*} \or\equal{#1}{<???>}
                     \or\equal{#1}{nn} \or\equal{#1}{ } \or\equal{#1}{}}
    { \begin{align*}%
         #2 %
         \end{align*}%
        }
    {\begin{equation}\label{#1}\begin{split}\allowdisplaybreaks%
         #2%
         \end{split}\end{equation}
        }
  }

\newcommand{\nit}{\mathbb{N}}
\def\N{\mbox{N}}            % Normal distribution
\renewcommand{\i}{\mbox{\bf i}}
\newcommand{\CO}{{\mathcal{O}}}
\def\Co{{\scriptstyle \mathcal{O}}} % small {\cal o}
\renewcommand{\O}{\CO}
\renewcommand{\o}{\Co}
\newcommand{\zit}{\mathbb{Z}}
\newcommand{\R}{\mathbb{R}}
\newcommand{\rr}{\mathcal{R}}
\newcommand{\tr}{\mathop{\rm{tr}}}
\newcommand{\ii}{\mathcal{I}}
\newcommand{\diag}{\mathop{\rm{diag}}}
\newcommand{\rank}{\mathop{\rm{rank}}}
\newcommand{\Var}{\mathop{\mbox{\sf Var}}}
\newcommand{\var}{\mathop{\mbox{\sf Var}}}
\newcommand{\Cov}{\mathop{\mbox{\sf Cov}}}
\newcommand{\cov}{\mathop{\mbox{\sf Cov}}}
\newcommand{\q}{\mathop{\mbox{Q}}}
\newcommand{\E}{\mathop{\mbox{\sf E}}}     % ISE
\renewcommand{\P}{\operatorname{P}}            % ISE
\newcommand{\Corr}{\mathop{\mbox{Corr}}}
\def\std{\mathop{\mbox{Std}}}            %standard deviation
\newcommand{\IF}{\boldsymbol{1}}    % indicator function
\def\defeq{\stackrel{\mathrm{def}}{=}}  % for definition
\def\eqdef{\stackrel{\rm def}{=}}
\def\eps{\ensuremath{\varepsilon}}
\def\iid{i.i.d.\xspace }
\def\argmax{\mathop{\mbox{arg\,max}}}
\def\argmin{\mathop{\mbox{arg\,min}}}
\def\ind{\IF}
\def\en{\infty}
\def\summ#1#2#3{\sum_{#1=#2}^{#3}}
\def\noframe{}
\def\t{^{\scriptstyle{\mathsf{T}}} \hspace{-0.15em}}
\newcommand{\scz}{{\cal Z}}

\newcommand{\alze}{{\alpha^{(0)}}}
\newcommand{\lleq}{\leqslant}
\newcommand{\ggeq}{\geqslant}
\newcommand{\talze}{{\widetilde{\alpha}^{(0)}}}
\newcommand{\zze}{{Z^{(0)}}}
\newcommand{\tzze}{{\widetilde{Z}^{(0)}}}
\newcommand{\hzt}{{\widehat{Z}_t}}
\newcommand{\zzet}{{Z_t^{(0)}}}
\newcommand{\tzzet}{{\widetilde{Z}_t^{(0)}}}
\newcommand{\zzetr}{{Z_t^{(0)\top}}}
\newcommand{\tzzetr}{{\widetilde{Z}_t^{(0)\top}}}

\newcommand{\hZ}{{\widehat{Z}}}
\newcommand{\hA}{{\widehat{A}}}
\newcommand{\ha}{{\widehat{\alpha}}}
\newcommand{\czzetr}{{\cZ_t^{(0)\top}}}
\newcommand{\chA}{{\widehat{\cA}}}
\newcommand{\chZ}{{\widehat{\cZ}}}
\newcommand{\cZ}{{\cal Z}}
\newcommand{\cA}{{\cal A}} 
\title{Dynamic Large Spatial Covariance Matrix Estimation in Application to Semiparametric Model Construction via Variable Clustering: the SCE approach}
\author{Song Song\thanks{University of California, Berkeley. Email: songsong@stat.berkeley.edu}}
\maketitle
\begin{abstract}
To better understand the spatial structure of large panels of economic and financial time series and provide a guideline for constructing semiparametric models, this paper first considers estimating a large spatial covariance matrix of the generalized $m$-dependent and $\beta$-mixing time series (with $J$ variables and $T$ observations) by hard thresholding regularization as long as ${{\log J \, \cx^*(\ct)}}/{T} = \Co(1)$ (the former scheme with some time dependence measure $\cx^*(\ct)$) or $\log J /{T} = \Co(1)$ (the latter scheme with the mixing coefficient $\beta_{mix}=\CO\{(J^{2+\delta'} \sqrt{\log J T})^{-1}\}, \delta' >0$. We quantify the interplay between the estimators' consistency rate and the time dependence level, discuss an intuitive resampling scheme for threshold selection, and also prove a general cross-validation result justifying this. Given a consistently estimated covariance (correlation) matrix, by utilizing its natural links with graphical models and semiparametrics, after ``screening'' the (explanatory) variables, we implement a novel forward (and backward) label permutation procedure to cluster the ``relevant'' variables and construct the corresponding semiparametric model, which is further estimated by the groupwise dimension reduction method with sign constraints. We call this the SCE (screen - cluster - estimate) approach for modeling high dimensional data with complex spatial structure. Finally we apply this method to study the spatial structure of large panels of economic and financial time series and find the proper semiparametric structure for estimating the consumer price index (CPI) to illustrate its superiority over the linear models.

{\it Keywords}: Time Series, Covariance Estimation, Regularization, Sparsity, Thresholding, Semiparametrics, Graphical Model, Variable Clustering

{\it JEL classification}: C13, C14, C32, E30, G10
\end{abstract}

\section{Introduction}\label{intro}
\subsection{Large Spatial Covariance Matrix}
Recent breakthroughs in technology have created an urgent need for high-dimensional
data analysis tools. Examples include economic and financial time series, genetic data, brain imaging, spectroscopic imaging, climate data and many others. To model high dimensional data, especially large panels of economic and financial time series as our focus here, it is very important to begin with understanding the ``spatial'' structure (over the space of variables instead of from a geographic point of view; also used in future for convenience) instead of simply assuming any specific type of parametric (e.g. linear) model first. Estimation of large spatial covariance matrix plays a fundamental role here since it can indicate a predictive relationship that can be exploited in practice. It is also very important in numerous other areas of economics and finance, including but not limited to handling heteroscedasticity of high dimensional econometric models, risk management of large portfolios, setting confidence intervals (or interval forecasts) on linear functions of the means of the components, variable grouping via graphs, dimension reduction by principal component analysis (PCA) and classification by linear or quadratic discriminant analysis (LDA and QDA). In recent years, many application areas where these tools are used have dealt with very high-dimensional datasets with relatively small sample size, e.g. the typically low frequency macroeconomic data.

It is well known by now that the empirical covariance matrix for samples of
size $T$ from a $J$-variate Gaussian distribution, $\N_J(\mu,\Sigma_J)$ is not a good estimator
of the population covariance if $J$ is large. If $J/T \rightarrow c \in (0, 1)$ and the covariance matrix $\Sigma_J=I$ (the identity), then the empirical distribution of the eigenvalues of the sample covariance
matrix $\hat \Sigma_J$ follow the Mar\^{c}enko-Pastur Law (\cite{0025-5734-1-4-A01}) and the eigenvalues are supported on $((1-\sqrt{c})^2, (1+\sqrt{c})^2)$. Thus, the larger $J/T$ is, the more spread out the eigenvalues are.

Therefore, alternative estimators for large covariance matrices have attracted a
lot of attention recently. Two broad classes of covariance estimators have emerged.
One is to remedy the sample covariance matrix and construct a better estimate by using approaches such
as banding, tapering and thresholding. The other is to reduce
dimensionality by imposing some structure on the data such as factor models, \cite{RePEc:eee:econom:v:147:y:2008:i:1:p:186-197}. Among the first class, regularizing the covariance matrix by banding or tapering relies on a natural ordering among variables and assumes that variables far apart in the ordering are only weakly correlated, \cite{wu:po:03}, \cite{Bickel08regularizedestimation}, \cite{ca:zh:11} among others. However, there are many applications, such as large panels of macroeconomic and financial time series, gene expression arrays and other spatial data, where there is no total ordering on the plane and no defined notion of distance among variables at all. These existing applications require estimators to be invariant under variable permutations such as regularizing the covariance matrix by thresholding, \cite{2009arXiv0901.3220E} and \cite{Bickel08regularizedestimation2}. In this paper, we consider thresholding of the sample spatial covariance matrix for high dimensional time series, which extends the existing work from the iid to the dependent scenarios. Under the time series setup, a very important question to ask is: how the time dependence will affect the estimate's consistency? This is the first question this paper is going to answer.

For time series, there have been two recent works by \cite{bi:ge:11} and \cite{xi:wu:11} about banding and tapering the large autocovariance matrices for univariate time series. But our goal here is to better understand the spatial structure of high dimensional time series, so we need a consistent estimate of the large spatial covariance matrix, especially under a mixture of serial correlation (temporal dynamics), high dimensional (spatial) dependence structure and moderate sample size (relative to dimensionality).

\subsection{Relation with Semiparametric Model Construction}
As mentioned at the very beginning, when the spatial structure of the high dimensional data (time series) is complex, instead of simply assuming any specific type of parametric (e.g. linear) model first, we could adopt the flexible nonparametric approach. Due to the ``curse of dimensionality'' disadvantage of full nonparametrics, various semiparametric models have been considered to maintain flexibility in modeling while attempting to deal with the ``curse of dimensionality'' problem. However, most of the prior semiparametric works were carried out under some (prefixed) specific classes of semiparametric models without discussing which ones might be closer to the actual data structure. More specifically, to model some dependent variable $y$ (or $x_J$) using explanatory variables $x_1, x_2, \ldots, x_{J-1}$ (very large $J-1$), they might suggest the following high dimensional single index (\cite{hu:ho:fe:10}) or additive models (\cite{me:va:bu:09}, \cite{ra:la:li:wa:09}) first and then perform various variable selection techniques to eliminate some $x$'s to avoid overfitting.
\begin{itemize}
\item $\E(y) =g(x_1\beta_1+x_2\beta_2+x_3\beta_3 + x_4 \beta_4+ \ldots + x_{J-1} \beta_{J-1})$, where $g$ is an unknown univariate link function, and $\beta_1, \beta_2, \beta_3, \ldots, \beta_{J-1}$ are unknown parameters that belong to the parameter space.

\item $\E(y)=g_1(x_1)+g_2(x_2)+g_3(x_3) + \ldots + g_{J-1}(x_{J-1})$, where $g_1, g_2, g_3, \ldots, g_{J-1}$ are the unknown functions to be estimated nonparametrically.
\end{itemize}

This approach encounters limitations from the following three perspectives. {\textit{First}}, when the dimensionality $J-1 \rightarrow \infty$, the prefixed assumption itself becomes more and more questionable. Is the single index model or the additive one closer to the actual data structure? Or maybe some other type of semiparametric structures is more suitable? We do not know. And this becomes more challenging when the sample size $T$ is small (with respect to dimensionality).

\textit{Second}, this - prefixing some specific semiparametric classes first and then selecting variables accordingly - approach is also challenged by another character of high dimensional economic and financial time series: strong spatial dependence (near-collinearity). Under near-collinearity, we
expect variable selection to be unstable and very sensitive to
minor perturbation of the data. In this sense, we do not expect
variable selection to provide results that lead to clearer economic
interpretation than principal components or ridge regression. This is actually due to the fact that although compared with the information criteria based $L_0$ and ridge regression type $L_2$ regularization methods, the Lasso type $L_1$ variable selection techniques (\cite{ti96}) could deal with large $J$ and require weaker assumptions on the design matrix $x$ (composed of $x_1, \ldots, x_{J-1}$), it still requires the following (as one of many similar requirements) \textit{restricted eigenvalue (RE)} assumptions from \cite{2008arXiv0801.1095B}: \textit{there exists a positive number $\kappa = \kappa(s)$ such that
\begin{eqnarray*}
\min \Big\{ \frac{|x^\T \Delta|_2}{ \sqrt{T}|\Delta_\rr |_2}: |\rr|\leqslant s, \Delta \in \mathbb{R}^{J-1} \backslash  \{0\}, \parallel \Delta_{\mathcal{R}^c}\parallel_{1} \leqslant 3 \parallel \Delta_{\rr}\parallel_{1}  \Big\} \geqslant \kappa,
\end{eqnarray*}
where $|\rr|$ denotes the cardinality of the set $\rr$, $\mathcal{R}^c$ denotes the complement of the set of indices $\rr$, and $\Delta_{\rr}$ denotes the vector formed by the coordinates of the vector $\Delta$ w.r.t. the index set $\rr$.} It is essentially a restriction on the eigenvalues of the Gram matrix $\Psi_T=x^\T x/T$ as a function of sparsity $s$. To see this, recall the definitions of \textit{restricted eigenvalue} and \textit{restricted correlation} in \cite{2008arXiv0801.1095B}:
\begin{align*}
\psi_{\min}(u)&= \min_{z\in \mathbb{R}^J-1: 1 \leqslant \mathcal{M}(z) \leqslant u}\frac{z^\T \Psi_T z}{|z|_2^2}, \quad 1\leqslant z \leqslant J-1,\\
\psi_{\max}(u)&= \max_{z\in \mathbb{R}^J-1: 1 \leqslant \mathcal{M}(z) \leqslant u} \frac{z^\T \Psi_T z}{|z|_2^2},  \quad 1\leqslant z \leqslant J-1,\\
\psi_{m_1, m_2} &=\max  \Big\{\frac{f_1^\T x_{I_1}^\T x_{I_2} f_2}{T|f_1|_2 |f_2|_2}: I_1 \bigcap I_2 = \varnothing, |I_i| \leqslant m_i, f_i \in \mathbb{R}^{I_i} \backslash \{0\}, i=1, 2\Big\},
\end{align*}
where $|I_i|$ denotes the cardinality of $I_i$ and $x_{I_i}$ is the $T \times |I_i|$ submatrix of $x$ obtained by removing from $x$ the columns that do not correspond to the indices in $I_i$. Lemma 4.1 in \cite{2008arXiv0801.1095B} shows that if the \textit{restricted eigenvalue} of the Gram matrix $\Psi_T$ satisfies $\psi_{\min}(2s) > 3 \psi_{s,2s}$ for some integer $1\leqslant s \leqslant (J-1)/2$, Assumption RE holds. Under this condition, the
Lasso type estimate's various oracle inequalities could be derived, e.g. \cite{2008arXiv0801.1095B}, where the upper bounds typically negatively depend on $\kappa$. From an economic point of view, this in fact requires that the dependence can not be too strong, which, unfortunately, is often unsatisfied for large panels of macroeconomic and financial data.

\textit{Third}, when the proposed high dimensional semiparametric model has a complex structure, finding a proper penalty term and the corresponding estimation method for variable selection in general might be very difficult, since, ideally, the penalty should depend not only on the coefficients, but also on the (shapes of the) \textit{unknown} nonparametric link functions. Several examples of regularizing high dimensional semiparametric models could be found in Chapter $5$ and $8$ of \cite{bu:va:11}, \cite{ra:la:li:wa:09} among others.

To this end, developing a \textit{specific model free} high dimensional spatial structure \textit{first} and then constructing the right class of semiparametric models seems important. Specifically speaking, given $x_1, x_2, \ldots, x_{J-1}$ and $y$ (or $x_J$), we try to find the index sets $\mathcal A_1, \mathcal A_2, \ldots, \mathcal A_S$ (possibly with overlapping elements) such that $y$ could be well approximated by:
\begin{equation}
\sum_{s=1}^S g_s\Big(  \sum_{l=1}^ {|A_s|}\beta_{sl}   x_{l \in \mathcal A_s}\Big)\defeq \sum_{s=1}^S g_s\Big(  \beta_s^\T x_{\mathcal A_s}\Big), \label{eq:semi}
\end{equation}
where
\begin{itemize}
\item $|\bf \cdot|$ denotes the cardinality of the set $\bf \cdot$; $S$ is the number of index sets $\mathcal A_1, \mathcal A_2, \ldots, \mathcal A_S$ and also the number of the \textit{unknown} univariate nonparametric link functions $g_1, \ldots, g_S$;
\item  $x_{\mathcal A_s}\defeq (x_l, \l \in \mathcal A_s)$ is a vector of regressors w.r.t. the index set $\mathcal A_s$; $\beta_{sl}, 1\leqslant s \leqslant S, 1\leqslant l \leqslant |\mathcal A_s|$ are the \textit{unknown} parameters in the parametric space; $\beta_s = (\beta_{s1}, \ldots, \beta_{s|\mathcal A_s|})$;
\item $\forall j \neq l, s \neq t$, $x_j \in \mathcal A_s$, $x_l \in \mathcal A_t$, $x_j$ and $x_l$ are (conditionally) independent given other $x$'s.
\end{itemize}
If $K \defeq |\mathcal A_1\bigcup \ldots \bigcup \mathcal A_S| \ll J-1$ (although $K$ is still possibly not moderate), we could strike a balance between dimension reduction and flexibility of modeling. Model \eqref{eq:semi} is very general and includes the single index model if $S=1$ (\cite{ic:93}), the additive model if $|\mathcal A_1|= |\mathcal A_2|= \ldots= |\mathcal A_S|=1$ (\cite{ha:ti:90}), the partial linear model if $S=2$, $g_1$ is the identity function and $|\mathcal A_2|=1$ (\cite{sp:88}) and the partial linear single index model if $S=2$ and $g_1$ is the identity function (\cite{an:po:93}, \cite{ca:fa:gi:wa:95}, \cite{yu:ru:02}). Model \eqref{eq:semi} could also be viewed as an extension of the multiple index model (\cite{st:86}, \cite{ic:le:91}, \cite{ho:98}, \cite{xi:08}) and can be further generalized if the RHS is $\mu \{\E (y)\}$ where $\mu$ is some known link function. As also considered by \cite{li:li:zh:10}, if $\mathcal A_1, \mathcal A_2, \ldots, \mathcal A_S$ are disjoint (no overlapping elements), then for each group of variables $x_{\mathcal A_s}$, we could say $g_s$ denotes (the only) one index. Thus according to \cite{li:li:zh:10}, model \eqref{eq:semi} is identifiable as every subspace of every group $x_{\mathcal A_s}$ is identifiable and could be solved efficiently by the grouping dimension reduction method in \cite{li:li:zh:10}, where they primarily assume that the grouping information is available. An immediate question is that given $x_1, \ldots, x_{J-1}$ ($J-1 \rightarrow \infty$), how can we extract $S$ groups of ``relevant'' $x$'s with corresponding index sets $\mathcal A_1, \ldots, \mathcal A_S$ and $|\mathcal A_1\bigcup \ldots \bigcup \mathcal A_S| \ll J-1$? This is the second question this paper is going to answer. From now on, we mainly study the case where $\mathcal A_1, \mathcal A_2, \ldots, \mathcal A_S$ are disjoint, although in Section \ref{permutation} we will also present the method generating overlapping index sets.

\begin{figure}
  \centering
  \includegraphics[width=4.5cm]{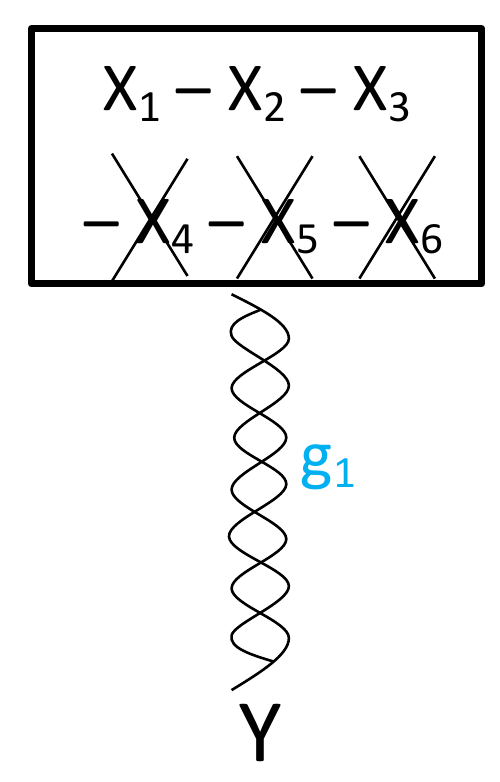}  \qquad \includegraphics[width=6cm]{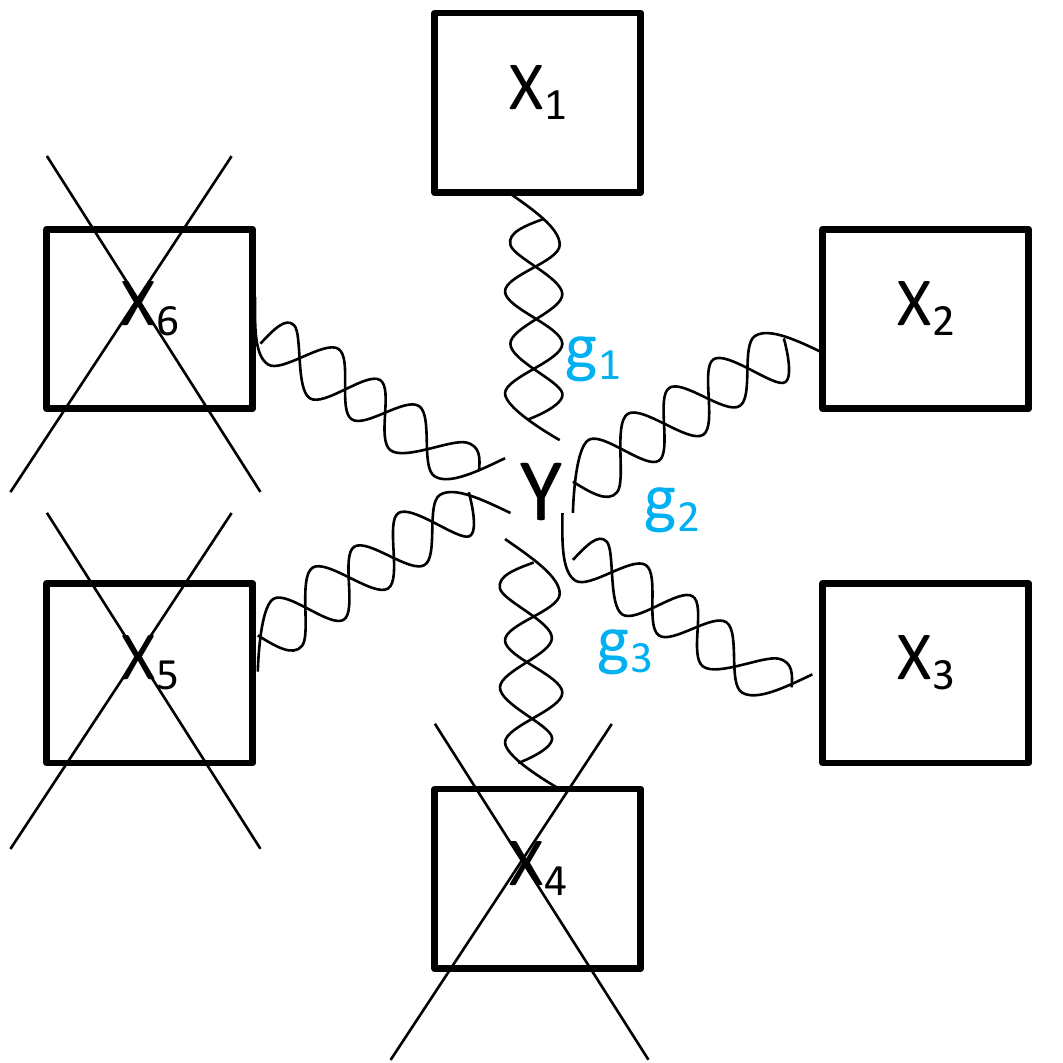}
  \caption{Left: $\E(y) =g(x_1\beta_1+x_2\beta_2+x_3\beta_3)$, where $g$ is an unknown univariate link function, and $\beta_1, \beta_2, \beta_3$ are unknown indices which belong
to the parameter space. Right: $\E(y) =g_1(x_1)+g_2(x_2)+g_3(x_3)$, where $g_1, g_2$ and $g_3$ are the unknown functions to be estimated nonparametrically.}\label{singleindex}
\end{figure}

\begin{figure}
\begin{center}
\includegraphics[width=6cm]{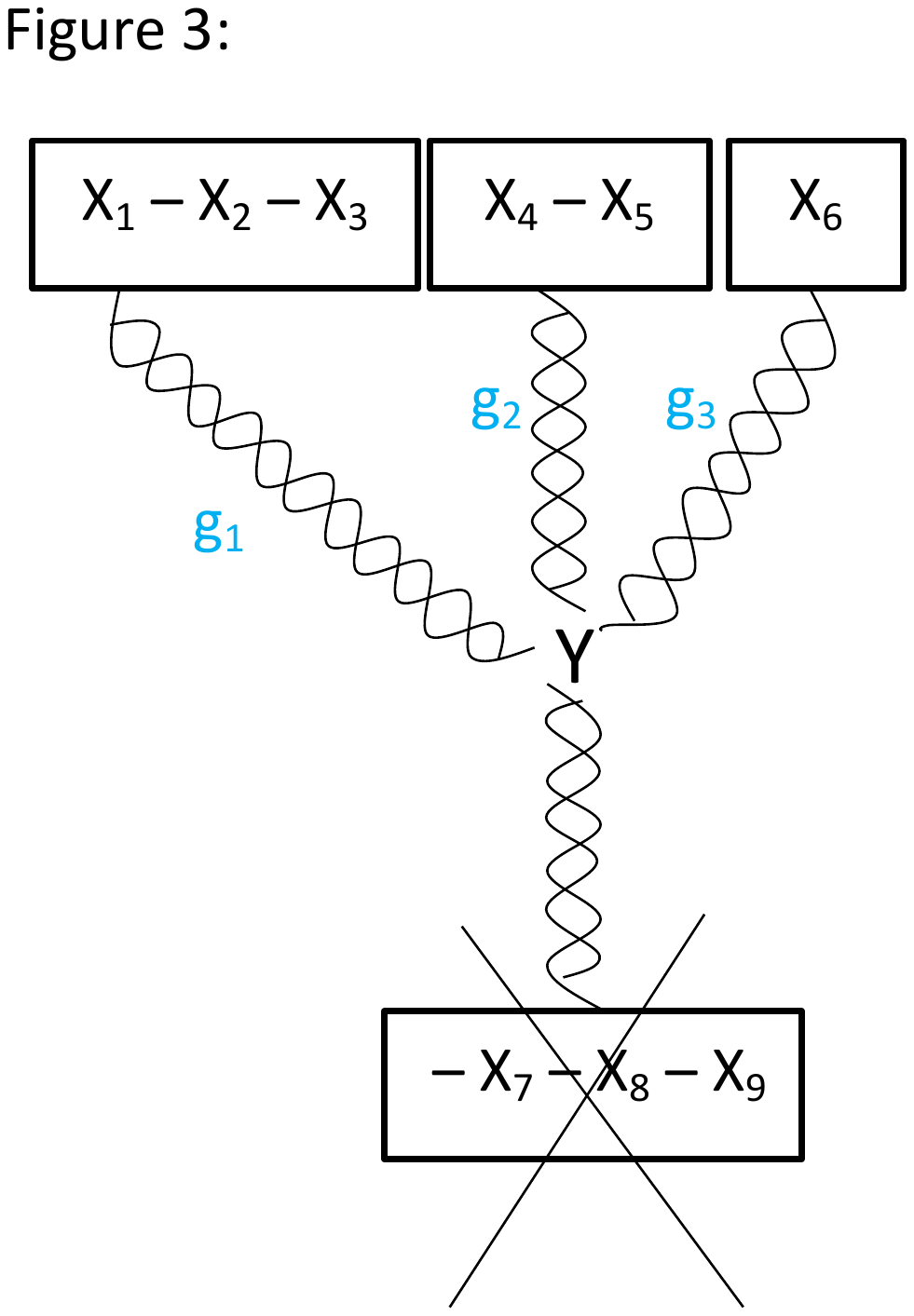}  \qquad
\includegraphics[width=7cm]{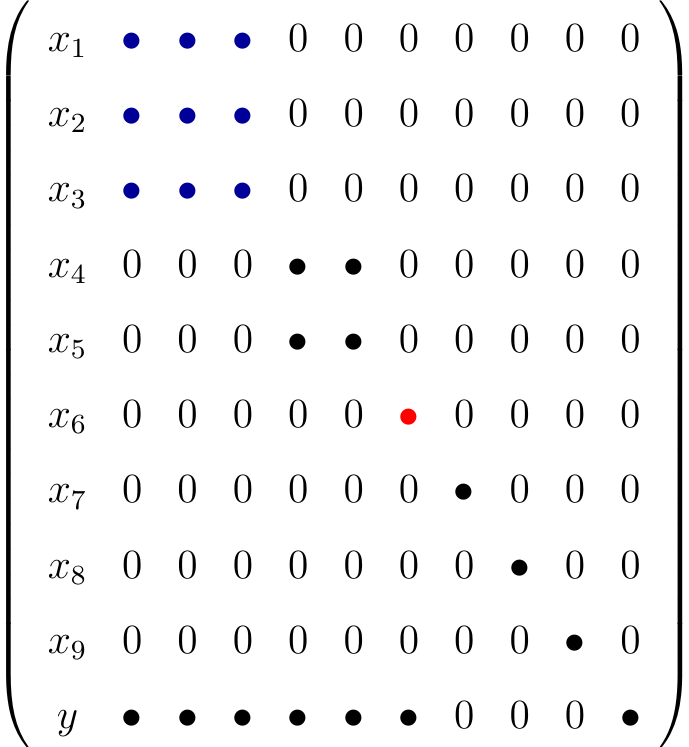}
%$\left(
%  \begin{array}{ccccccccccc}
%{x_1} &{\color{iseblue}\bullet} & {\color{iseblue}\bullet} & {\color{iseblue}\bullet} & 0 & 0 & 0 & 0 & 0& 0& 0\\
%x_2 &{\color{iseblue}\bullet} & {\color{iseblue}\bullet} & {\color{iseblue}\bullet} & 0 & 0 & 0 & 0 & 0& 0& 0\\
%x_3 & {\color{iseblue}\bullet} & {\color{iseblue}\bullet} & {\color{iseblue}\bullet} & 0 & 0 & 0& 0 & 0 & 0& 0\\
%x_4&             0 & 0 & 0 & \bullet & \bullet & 0 & 0& 0& 0& 0\\
%     x_5 &      0 & 0 & 0 & \bullet & \bullet & 0 & 0& 0& 0& 0\\
%                x_6 &  0 & 0 & 0 & 0 & 0 & {\color{isered}\bullet} & 0& 0& 0& 0\\
%                x_7  & 0 & 0 & 0& 0 & 0 & 0 & {\bullet} & 0& 0& 0\\
%               x_8   & 0 & 0 & 0& 0 & 0 & 0 & 0 & {\bullet}& 0& 0\\
%                          x_9   &  0 & 0 & 0& 0 & 0 & 0 & 0 & 0& {\bullet}& 0\\
%                          y &{\bullet} & {\bullet} & {\bullet} & {\bullet} & {\bullet} & {\bullet} & 0 & 0& 0& {\bullet}\\
%  \end{array}
%\right)$
\end{center}
  \caption{Left: $\E(y) =g_1(x_1\beta_1+x_2\beta_2+x_3\beta_3)+g_2(x_4\beta_4+x_5\beta_5)+g_3(x_6\beta_6)$, where $g_1, g_2$ and $g_3$ are the unknown functions to be estimated nonparametrically. Right: a sample of block diagonal structure after label permutation to the regularized large spatial covariance matrix. }\label{multipleindex}
\end{figure}

Before moving on, let us study the differences among various semiparametric models from the graphical point of view. If we use a vertex in the graph to represent a relevant variable, a solid edge in a ``block'' to represent linear relationship among variables inside, a bandy edge (connecting a ``block'' with the dependent variable $y$) to represent a nonparametric link function, a crossed vertex to represent an ``unrelated'' ones, then we can visualize different semiparametric models through corresponding graphs. For instance, we can get Figure \ref{singleindex} (left) for the single index model; Figure \ref{singleindex} (right) for the additive model; Figure \ref{multipleindex} (left) for the (more general) multiple index model, among many others. As we can see, the underlying difference among various semiparametric models is where to allocate the nonparametric link function and linearity through clustering variables. Consequently, assuming that all the variables have been included (complete graph), if we can find the corresponding type of graphs, we can construct the right class of semiparametric models. Sparse concentration matrices are of special interest in graphical models because zero partial correlations help establish independence and conditional independence relations in the context of graphical models and thus imply a graphical structure. For example, if we have a sparse covariance matrix for $y, x_1, \ldots, x_9$ as the one in Figure \ref{multipleindex} (right), we know that $x_1, \ldots, x_6$ are ``relevant'' to $y$, and due to the ``block'' structure w.r.t. $x_1, x_2, x_3$ and $x_4, x_5$, we can construct the following class of semiparametric models as a specific case of \eqref{eq:semi}:
\begin{equation}
\E (y) =g_1(x_1\beta_1+x_2\beta_2+x_3\beta_3)+g_2(x_4\beta_4+x_5\beta_5)+g_3(x_6\beta_6). \label{eq:multipleindex} \end{equation}
Now we have found the links among semiparametrics, graphical models and sparse large spatial covariance matrix. Thus consistently estimating the large sparse covariance matrix first and clustering the (explanatory) variables (or forming a block diagonal structure for the corresponding partition of the covariance matrix) are the key focuses. In this article, we assume that the grouping structure (or the corresponding covariance matrix) and parametric coefficients $\beta$s are both time invariant to simply the study.

Another related and potential application of clustering variables comes from group regularization (e.g. group Lasso, \cite{RePEc:bla:jorssb:v:68:y:2006:i:1:p:49-67}) in the modern sparsity analysis. \cite{hu:zh:09} show that, if the underlying structure is strongly group-sparse, group Lasso is more robust to noise due to the stability associated with group structure and thus requires a smaller sample size to meet the sparse eigenvalue condition required in modern sparsity analysis. However, other than the situations, e.g. multi-task learning, where we have clear background knowledge about how to group variables, in general, it is hard to tell how to properly group the variables to make use of group regularization. An example could be found in a paper in preparation with Bickel, \cite{s2:bi:10}, where they discuss three types of estimates for large vector auto regression w.r.t. different grouping methods. To this end, proper ``grouping'' of the variables is also significant.

For the semiparametric modeling in econometrics, people usually ``group'' the variables in a ``rule of thumb'' way. For example, to model the consumer price index (CPI - all items), they might subjectively group the variables ``CPI - apparel \& upkeep; transportation; medical care; commodities; durables; services'' in the first group, ``CPI - all items less food; all items less shelter; all items less medical care'' in the second group; `` Producer Price Index (PPI) - Finished Goods; Finished Consumer Goods; Intermed Mat. Supplies \& Components; Crude Materials'' in the third group, ``Implicit Price Deflator (of Personal Consumption
Expenditures) PCE - all items; durables; nondurables; services'' in the fourth group, all other variables in the last group. Is this way of grouping closest to the actual data structure? Why not put ``CPI - Durables; PCE - Durables'' in one group and ``CPI - Services; PCE - Services'' in another group? We are going to provide a procedure of grouping these variables from a data-driving approach.

In summary, the novelty of this article lies in the following two aspects. {\textit{First}}, under the high dimensional time series situation, we show consistency (and the explicit rate of convergence) of the threshold estimator in the operator norm, uniformly over the class of matrices that satisfy our notion of sparsity as long as ${{\log J \, \cx^*(\ct)}}/{T} = \Co(1)$ (for the generalized $m$-dependent time series; the meaning of $\cx^*(\ct)$ is presented later) or $\log J /{T} = \Co(1)$ (for the $\beta$-mixing process with the mixing coefficient $\beta_{mix}=\CO\{(J^{2+\delta'} \sqrt{\log J T})^{-1}\}, \delta' >0$. Furthermore, we quantify the interplay between the estimators' consistency rate and the time dependence level, which is novel in this context. There are various arguments showing that convergence in the operator norm implies convergence of eigenvalues of eigenvectors, \cite{2009arXiv0901.3220E} and \cite{Bickel08regularizedestimation2}, so this norm is particularly appropriate for
various applications. We also discuss an intuitive resampling scheme for threshold selection for high dimensional time series, and prove a general cross-validation result that justifies this approach. {\textit{Second}}, we propose a SCE (screen - cluster - estimate) approach for modeling high dimensional data with complex spatial structure. Specifically, given a consistently estimated large spatial covariance (correlation) matrix, by utilizing its natural links with graphical models and semiparametrics and using the correlation (or covariance for the standardized observations) between variables as a measure of similarity, after ``screening'' the (explanatory) variables, we propose a novel forward (and backward) label permutation procedure to cluster the ``relevant'' (explanatory) variables (or to form a block diagonal structure for the regularized large spatial matrix) and construct the corresponding semiparametric model, which is further estimated by the groupwise dimension reduction method (\cite{li:li:zh:10}) with sign constraints.

It is noteworthy that the ``screening'' in Step $1$, ``clustering'' in Step $2$, and the ``sign constraints'' in Step $3$ here are crucial for applying the groupwise dimension reduction method of \cite{li:li:zh:10} in the high dimensional situation. \textit{First}, their method requires the use of the high dimensional kernel function, which faces some limitations when $J \gg T$. The Step $1$ here help reduce the dimensionality from $J$ ($\gg T$) to a more manageable level. \textit{Second}, they primarily assume that the grouping information is available from the background knowledge, which is often not available from the typically (spatially) unordered high dimensional data sets. Although they also proposed an information criterion based grouping method, this - ``trying'' many different combinations of grouping - approach is very computationally intensive and less practical. The Step $2$ here provides this grouping information from a data driven approach with feasible computation. \textit{Third}, without adding the ``sign constraints'', the signs of the estimated parametric coefficients might violate the economic laws (details presented in Section \ref{application}). Overall, together with \cite{li:li:zh:10}'s very timely and stimulating work, we provide an \textit{integrated} approach for modeling high dimensional data with complex spatial structure.

The rest of the article is organized as follows. In the next section, we present the main notations of the thresholding estimator. The estimates' properties are presented in Section \ref{asymptotic}. In Section \ref{permutation} we state the details of the SCE procedure and in Section \ref{application} apply it to study the spatial structure of large panels of macroeconomic and financial times series and find the proper semiparametric structure for estimating the consumer price index (CPI). Section \ref{discussion} contains concluding remarks with a brief discussion. All technical proofs are sketched in the appendix.

\section{Dynamic Large Spatial Covariance Matrix Estimation}\label{metho}
We start by setting up notations and corresponding concepts for covariance matrix $\Sigma$, which are mostly from \cite{Bickel08regularizedestimation} and \cite{Bickel08regularizedestimation2}. We write $\lambda_{\max}(\Sigma)=\lambda_1(\Sigma) \geqslant \ldots \geqslant \lambda_J(\Sigma)=\lambda_{\min}(\Sigma)$ for the eigenvalues of a matrix $\Sigma$. Following the notations of \cite{Bickel08regularizedestimation} and \cite{Bickel08regularizedestimation2}, we define that, for any $0 \leqslant r, s \leqslant \infty$ and a $J \times J$ matrix $\Sigma$, $\|\Sigma\|_{(r,s)} \defeq \sup\{\|\Sigma x\|_s: \|x\|_r =1\},$ where $\|x\|_r^r=\sum_{j=1}^J |x_j|^r$. In particular, we write $\|\Sigma\|=\|\Sigma\|_{(2,2)}=\max_{1\leqslant j \leqslant J}|\lambda_j(\Sigma)|$, which is the operator norm for a symmetric matrix. We also use the Frobenius matrix norm, $\|\Sigma\|_F^2=\sum_{i,j}\sigma_{ij}^2=tr(\Sigma \Sigma^\T)$. Dividing it by a factor $J$ brings $\|\Sigma\|_F^2/J$, which is the average of a set of eigenvalues, while the operator norm $\|\Sigma\|_{(2,2)}$ means the maximum of the same set of eigenvalues. \cite{Bickel08regularizedestimation2} defines the thresholding operator by
$$ T_s(\Sigma)\defeq [m_{ij} \IF(|m_{ij}|\geqslant s)],$$
which we refer to as $\Sigma$ thresholded at $s$. Notice that $T_s$ preserves symmetry; it is invariant under permutations of variable labels; and if $\|T_s - T_0\|\leqslant \eps$ and $\lambda_{\min}(\Sigma)> \eps$, it preserves positive definiteness.

We study the properties of the following uniformity class of covariance matrices invariant under permutations
$$\mathcal{U}_\tau(q,c_0(J),M) \defeq\{\Sigma: \sigma_{ii}\leqslant M, \sum_{j=1}^J |\sigma_{ij}|^q \leqslant c_0(J), \forall i \}, \;0 \leqslant q <1.$$

We will mainly write $c_0$ for $c_0(J)$ in the future. Suppose that we observe $T$ $J$-dimensional observations $X_1, \ldots, X_T$ with $\E X=0$ (without loss of generality), and $\E (X X^\T)=\Sigma$, which is independent of $t$. We consider the sample covariance matrix by
\begin{equation} \label{def:sample}
\hat \Sigma \defeq T^{-1} \summ tiT (X_t - \bar X)(X_t - \bar X)^\T \defeq [\hat \sigma_{ij}],
\end{equation}
with $\bar X= T^{-1} \sum_{t=1}^T X_t$.

Let us first recall the fractional cover theory based definition, which was introduced by \cite{ja:04} and can be viewed as a generalization of $m$-dependency.
Given a set $\mathcal{T}$ and random variables $V_t$, $t\in \mathcal{T}$, we say:
\begin{itemize}
\item A subset $\mathcal{T}'$ of $\ct$ is $independent$ if the corresponding random variables $\{V_t\}_{t\in \ct'}$ are independent.
\item A family $\{\ct_j\}_j$ of subsets of $\ct$ is a $cover$ of $\ct$ if $\bigcup_j \ct_j = \ct$.
\item A family $\{(\ct_j, w_j)\}_j$ of pairs $(\ct_j, w_j)$, where $\ct_j \subseteq \ct$ and $w_j \in [0, 1]$ is a {\it fractional cover} of $\ct$ if $\sum_j w_j \IF_{\ct_j} \geqslant \IF_{\ct}$, i.e. $\sum_{j: t\in \ct_j} w_j \geqslant 1$ for each $t \in \ct$.
\item A (fractional) cover is $proper$ if each set $\ct_j$ in it is independent.
\item $\cx(\ct)$ is the size of the smallest proper cover of $\ct$, i.e. the smallest $m$ such that $\ct$ is the union of $m$ independent subsets.
\item $\cx^*(\ct)$ is the minimum of $\sum_j w_j$ over all proper fractional covers $\{(\ct_j, w_j)\}_j$.
\end{itemize}
Notice that, in spirit of these notations, $\cx(\ct)$ and $\cx^*(\ct)$ depend not only on $\ct$ but also on the family $\{V_t\}_{t \in \ct}$. Further note that $\cx^*(\ct) \geqslant 1$ (unless $\ct =\varnothing$) and that $\cx^*(\ct) =1 $ if and only if the variables $V_t, t\in \ct$ are independent, i.e. $\cx^*(\ct)$ is a measure of the dependence structure of $\{V_t\}_{t \in \ct}$. For example, if $V_t$ only depends on $V_{t-1}, \ldots, V_{t-k}$ but is independent of all $\{V_s\}_{s< t-k}$, we will have $k+1$ independent sets:
\begin{align*}
\ct_1 &=\{V_1, V_{(k+1)+1}, V_{2(k+1)+1}, \ldots\},\\
\ct_2 &=\{V_2, V_{(k+1)+2}, V_{2(k+1)+2}, \ldots\},\\
& \ldots\\
\ct_{k+1} &=\{V_{k+1}, V_{(k+1)+(k+1)}, V_{2(k+1)+(k+1)}, \ldots\},
\end{align*}
s.t. $\bigcup_{j=1}^{k+1} \ct_j = \ct$. So $\cx^*(\ct) =k+1$ (if $k+1 < T$).

Besides the generalized $m$-dependent process, we are also going to consider the $\beta$-mixing process, which is related to the underlying measures of dependence between $\sigma$-fields. More precisely, let $(\Omega, \mathcal{A}, \P)$ be a probability space and $\mathcal{U}, \mathcal{V}$ be two sub $\sigma$-algebras of $\mathcal{A}$, the $\beta$-mixing coefficient $\beta(\mathcal{U}, \mathcal{V})=\E esssup \{|\P(V/\mathcal{U}-\P(V))|; V \in \mathcal{V}\}$ be a measure of dependence between $\mathcal{U}$ and $\mathcal{V}$, which has been defined by Kolmogorov and first appeared in the paper by \cite{volkonskii:178}. By its definition, the closer to $0$ $\beta$ is, the more independent the time series is. For examples of the $\beta$-mixing process, we refer to \cite{do:94}. Through this article, we use $\beta_{mix}$ to denote the $\beta$-mixing coefficient for notational convenience.

\section{Estimates' Properties}\label{asymptotic}
We have the following two results which parallel those in \cite{Bickel08regularizedestimation} and \cite{Bickel08regularizedestimation2}.
\subsection{Interplay Between Consistency Rate and Time Dependence Level}
\begin{theorem}[Dependence level affects consistency?] \label{dependent}
Suppose for all $i, j$, $|X_{ti}X_{tj}| \defeq |V_t| \leqslant M_t$ holds with a high probability and $\sum_{t=1}^T M_t^2/T$ is bounded by some constant $C'$. Then, uniformly on $\mathcal{U}_{\tau} (q,c_0(J),M)$, for sufficiently large $M'$ also depending on $C'$, if
$$s_T= M'(C') \sqrt{\frac{{\log J \, \cx^*(\ct)}}{T}}$$
and ${{\log J \, \cx^*(\ct)}}/{T} = \Co(1)$, then
\begin{align*}
\|T_{s_T}(\hat \Sigma) - \Sigma\|&=\CO_P \Big[c_0(J)\Big\{\frac{{\log J \, \cx^*(\ct)}}{T} \Big\}^{(1-q)/2}\Big]\\
J^{-1}\|T_{s_T}(\hat \Sigma) - \Sigma\|_F^2&=\CO_P \Big[c_0(J)\Big\{\frac{{\log J \, \cx^*(\ct)}}{T}\Big\}^{1-q/2}\Big]
\end{align*}
\end{theorem}
Not surprisingly, this theorem states that if we use the hard thresholding method to regularize the large sample covariance matrices, the consistency rate gets slower when the dependence level ($\cx^*(\ct)$) increases, or in other words, the rate is maximized when $\cx^*(\ct)=1$, same as what \cite{Bickel08regularizedestimation2} shows for the i.i.d case. When $\cx^*(\ct)$ reaches $T$, it will be offset by $T$ in the denominator. The intuition behind is clear: if dependence is strong, then additional information brought by a ``new'' observation will be effectively less, i.e. the overall information from $T$ observations will be less correspondingly, which will result in a slower consistency rate. On the other hand, according to the ${{\log J \, \cx^*(\ct)}}/{T} = \Co(1)$ requirement, when the dependence level $\cx^*(\ct)$ increases, $J$ must decrease and $T$ must increase to retain the same amount of information.

A very natural question to ask next is: to what extent, the degree of dependence (in terms of $\beta$-mixing coefficients) is allowed, while the consistency rate is still the same as the i.i.d. case, i.e. to study the relationship among high dimensionality $R$, moderate sample size $T$ and $\beta$-mixing coefficient $\beta_{mix}$.

\begin{assumption}\label{ass:mixing}
\begin{itemize}
\item[A1] $\forall t$, $\E X_{ti}X_{tj}=0$
\item[A2] $\exists \sigma^2$, $\forall n, m$, $m^{-1}\E(X_{ni}X_{nj}+\ldots+X_{n+m, i}X_{n+m, j})^2 \leqslant \sigma^2$
\item[A3] $\forall t$, $|X_{ti}X_{tj}| \leqslant M$
\end{itemize}
\end{assumption}

\begin{theorem}[Balance ``$J, T, \beta$'' to achieve ``good'' consistency rate]
\label{theo:mixing}
Assume the $\beta$-mixing sequence $\{X_{ti}X_{tj}\}_{t=1}^T$ satisfies Assumption \ref{ass:mixing} $\forall i, j$ with a high probability. Then, uniformly on $\mathcal{U}_{\tau} (q,c_0(J),\Sigma)$, for sufficiently large $M'$ also depending on $\sigma^2, M$, if
$s_T= M'(\sigma^2, M) \sqrt{\frac{\log J}{T}}$, $\log J / T = \Co(1)$ and the $\beta$-mixing coefficient {$\beta_{mix}=\CO\{(J^{2+\delta'} \sqrt{\log J T})^{-1}\}, \delta' >0$}, we have:
\begin{align*}
\|T_{s_T}(\hat \Sigma) - \Sigma\|&=\CO_P \Big\{c_0(J)\Big(\frac{\log J}{T}\Big)^{(1-q)/2}\Big\}\\
J^{-1}\|T_{s_T}(\hat \Sigma) - \Sigma\|_F^2&=\CO_P \Big\{c_0(J)\Big(\frac{\log J}{T}\Big)^{1-q/2}\Big\}
\end{align*}
\end{theorem}
As we can see, when dimensionality $J$ increases, since the $\beta$-mixing coefficient is controlled by $\CO\{(J^{2+\delta'} \sqrt{\log J T})^{-1}\}, \delta' >0$, the dependence level must decrease at the rate of $J^{-2}$ (skipping the slow varying $\log$s). When $J$ is very large, this means ``nearly'' independent, which again confirms the result from the previous theorem.

\subsection{Choice of Threshold via Cross Validation} \label{sub:threscv}
\begin{figure}
\centering
  \includegraphics[width=14cm]{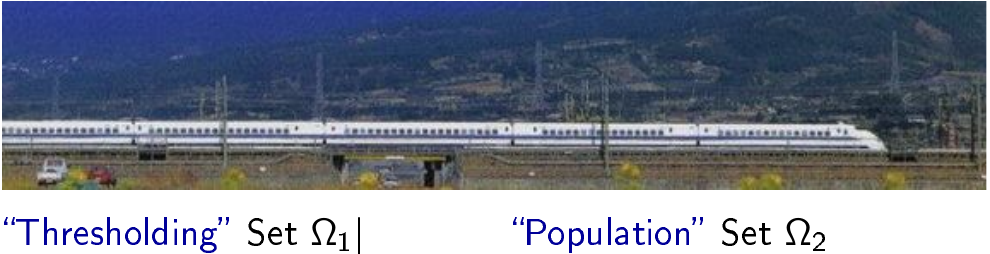}
  \caption{Illustration of the cross-validation method.} \label{cv}
    \end{figure}
Choices of threshold play a fundamental role in implementing this estimation procedure. We choose an optimal threshold by a cross-validation procedure as in \cite{Bickel08regularizedestimation2} and \cite{Bickel08regularizedestimation}. In particular, we divide the data set $\Omega$ of size $T$ into two consecutive segments, $\Omega_1$ and $\Omega_2$ of size $T_1$ and $T_2$ respectively, where $T_1$ is typically about $T/3$. Then we compare the regularized (via thresholding) ``target'' quantity $T_s(\hat \Sigma_{1,v})$, estimated from $\Omega_1$, with the ``target'' quantity $\hat \Sigma_{2,v}$, estimated from $\Omega_2$. Hence $\hat \Sigma_{2,v}$ can be viewed as a proxy to the population ``target'' quantity $\Sigma$. The subindex $v$ in $T_s(\hat \Sigma_{1,v})$ and $\hat \Sigma_{2,v}$ indicates values from the $v$th split from a total of $N$ repeats. The optimal threshold is then selected as a minimizer (w.r.t. $s$) of the empirical loss function over $N$ repeats, i.e.
\begin{equation}
\argmin_s N^{-1} \sum_{v=1}^N\|T_s(\hat \Sigma_{1,v}) - \hat \Sigma_{2,v}\|_F^2. \label{empirical}
\end{equation}
Similarly, the oracle threshold is then selected as a minimizer w.r.t. $s$ of the oracle loss function over $N$ repeats, i.e.
\begin{equation}
\argmin_s \E \|T_s(\hat \Sigma_{1,v}) - \hat \Sigma_{2,v}\|_F^2. \label{oracle}
\end{equation}

Since the data are observed in time, the order of $X_t$ is of importance, and hence a random split of $\Omega$ to $\Omega_1$ and $\Omega_2$ is not appropriate in a time series context. Alternatively, we randomly select a consecutive segment of size $T_1 + T_2$ as $\Omega_1 \bigcup \Omega_2$ from the data set $\Omega$ first, and then take the first third of $\Omega_1 \bigcup \Omega_2$ as $\Omega_1$ ($T_2 \approx 2 T_1$) and the remaining two thirds as $\Omega_2$. Figure \ref{cv} provides an illustration for the cross-validation procedure. We repeat this $N$ times as before. Our goal now is to show that the rates of convergence for the empirical loss function and the oracle loss function are of the same order and hence, asymptotically the empirical threshold $\hat s$ performs as well as the oracle threshold $s_0$ selection.

Our theoretical justification is based on adapting the results on the optimal threshold selection in \cite{Bickel08regularizedestimation2} and the optimal band selection in \cite{bi:ge:11} to a case of optimal choice of a threshold for high dimensional $\beta$-mixing time series. Let $W_1, \ldots, W_n, \ldots, W_{n+B}$ be $J^2 \times 1$ vectors with common mean $\E W$. Let $\|x\|_v = \max_{p=1, \ldots, P}|v'_p x|, \; x \in \R^J, v_p \in \R^J, \|v_p\|=1$ and $\bar{W_B}=B^{-1} \sum_{p=1}^B W_{n+p}$. Then the empirical and oracle estimates based on $W_k$ are defined as
\begin{align}
\hat \mu^e &\defeq \argmin_{p=1, \ldots, P} |\bar{W_B}-\hat \mu_p|^2 \label{emploss}\\
\hat \mu^o &\defeq \argmin_{p=1, \ldots, P} |\E W-\hat \mu_p|^2 \label{oraloss}
\end{align}
respectively, where $\hat \mu_p$ is estimated using $W_1, \ldots, W_n$.

We use Theorem $3$ in \cite{Bickel08regularizedestimation2} as Lemma \ref{cvpreparation} here, which states a result on asymptotic relation between the empirical and oracle estimates $\hat \mu^e$ and $\hat \mu^o$.
\begin{lemma} [Theorem 3 in \cite{Bickel08regularizedestimation2}]
\label{cvpreparation}
If the following assumptions (A4, A5, A6) are satisfied
\begin{itemize}
\item[A4] $|\hat \mu^o - \E W|^2=\Omega_P(r_n)$;
\item[A5] $\E \max_{p=1, \ldots, P}\|(v_j, W_1-\mu)\|^2 \leqslant C \rho(P)$ for $v_p \in \R^J, \|v_p\|=1$;
\item[A6] $\rho(P_n)=\Co(r_n)$,
\end{itemize}
then we have
$$|\hat \mu^e - \bar W_B|^2=|\hat \mu^o-\E W|^2\{1+\Co(1)\}=\Omega_P(r_n).$$
\end{lemma}
Without loss of generality, assume that the number of repeats $N=1$. Notice that the empirical estimates $T_s(\hat \Sigma_{1,v})$ and $\hat \Sigma_{2,v}$ play the role of $\hat \mu_p$ and $\bar W$ here respectively. Hence, if we can verify the conditions of Lemma \ref{cvpreparation}, we can apply it to justify the choice of a threshold by cross-validation and show that such regularized covariance matrix of high dimensional time series $\{X_t\}$, with an empirical selected threshold, asymptotically coincides with the regularized estimate selected by oracle. To this end, we also need the auxiliary Lemma \ref{covlemma3}.

\begin{lemma}
\label{covlemma3}
Assume that $v_t$ is white noise satisfying $\E v_t=0, \E v_t^2=\sigma^2$ and $\E |v_t|^\beta \leqslant C < \infty$ for $\beta>2$. Let $\|V\|_F=1$. For the $\beta$-mixing process satisfying the conditions of Theorem \ref{theo:mixing}, we have
\begin{align*}
\P\Big(J^{-1}|tr(V {\hat \Sigma_B} - V\Sigma)|\geqslant s \Big) &\leqslant K_1 \exp(-K_2 s^2 B)\\
J^{-1}\E \max_{p=1, \ldots, P} \Big(|tr\{v_j {\hat \Sigma_B} - \E (v_j\Sigma)\}| \Big) &\leqslant C(q, c_0, M)\sqrt{{\log P}/{B}}
\end{align*}
with some constants $K_1$ and $K_2$.
\end{lemma}

\begin{theorem}[Consistency of Cross Validation] \label{th:cv}
Let $\hat s$ and $s^o$ be the threshold selected from minimizing the empirical and oracle loss functions \eqref{empirical} and \eqref{oracle} respectively. Then under the conditions of Theorem \ref{theo:mixing} and $\CO_P=\Omega_P$, if $B_T = T \varepsilon(T, J)$, $\log P=\Co\{T^{q/2}c_0(J)J^{-1}(\log J)^{1-q/2}{\eps(T, J)}\}$, based on Lemma \ref{cvpreparation} and \ref{covlemma3}, then
\begin{align*}
\|T_{\hat s}(\hat \Sigma) - \Sigma\|_F = \|T_{s^o}(\hat \Sigma) - \Sigma\|_F\{1+\Co_P(1)\}.
\end{align*}
\end{theorem}
%\begin{itemize}
%\item $B_n$ ``training'' set sample size, $\hat s = m'\sqrt{\frac{\log J}{T}}: 1 \leqslant m' \leqslant M'$
%\end{itemize}

\section{The Screen - Cluster - Estimate (SCE) Procedure}\label{permutation}
To circumvent the problems in semiparametric modeling for high dimensional data with complex spatial structure, in the following three subsections, we state the three-step SCE procedure for constructing and estimating semiparametric models from a large number of unordered explanatory variables with a moderate sample size.

\subsection{Screen}
\begin{itemize}
\item[1] Estimate the $J \times J$ (dependent variable $y$ ($x_J$) and all explanatory variables $x_1, \ldots, x_{J-1}$) large covariance (Spearman's correlation) matrix using hard thresholding as $T_{\hat s}(\hat \Sigma) \defeq [\tilde \sigma_{ij}]$, and only keep and consider the (say $K$) $x$'s with nonzero correlation entries with $y$ for following steps. Without loss of generality, we rename the $K$ $x$'s as $x_1, x_2, \ldots, x_K$.
\end{itemize}

Since all observations are standardized first, the previously considered covariance matrix is actually the (Pearson's) correlation coefficient matrix. However, at the ``screening'' step, we estimate and threshold the large Spearman's rank correlation matrix, where Spearman's rank correlation between $x_i$ and $x_j$ is defined as:
\begin{equation}
\rho_{x_i, x_j}=\frac{\Cov\{F_i(x_i), F_j(x_j)\}}{\sqrt{\Var\{F_i(x_i)\}\Var\{F_j(x_j)\}}}, \label{def:rank}
\end{equation}
and $F_i$ and $F_j$ are the cumulative distribution functions of $x_i$ and $x_j$ respectively. It can be seen that the population version of Spearman's rank correlation is just the classic Person's correlation between $F_i(x_i)$ and $F_j(x_j)$. Here we consider Spearman's rank correlation instead of the Pearson's correlation coefficient is because the latter one is sensitive only to a linear relationship between two variables, while the former one is more robust than the Pearson's correlation - that is, more sensitive to nonlinear relationships. It could be viewed as a non-parametric measure of correlation and especially suitable for the non and semiparametric situations we consider here. It assesses how well an arbitrary monotonic function could describe the relationship between two variables. Specifically speaking, it measures the extent to which, as one variable increases, the other variable tends to increase, without requiring that increase to be represented by a linear relationship. If, as the one variable increases, the other decreases, the rank correlation coefficients will be negative. Similar to the consistency results towards the large spatial thresholding covariance (correlation) matrix studied here, \cite{xu:bi:10} established those for the large Spearman's rank correlation matrix (for the i.i.d. case).

 At step $1$, via hard thresholding, we single out the important predictors by using their Spearman's rank correlations with the response variable $y$ and eliminate all explanatory variables that are ``irrelevant'' to $y$.
In light of equation \eqref{eq:semi}, we actually get an estimate for $\mathcal A_1\bigcup \ldots \bigcup \mathcal A_S$. Thus we could reduce the feature space significantly from $J$ to a lower dimensional and more manageable space. Correlation learning is a specific case of independent learning, which ranks the features according to the marginal utility of each feature. The computational expediency and stability are prominently featured in independent learning. This kind of idea is frequently used in applications (\cite{gu:el:03}) and recently has been carefully studied for its theoretical properties by \cite{fa:lv:08} using Pearson's correlation for variable screening of linear models; \cite{hu:ho:sh:08} , who proposed the use of marginal bridge estimators to select variables for sparse high dimensional regression models; \cite{fa:fe:so:11} using the marginal strength of the marginal nonparametric regression for variable screening of additive models; \cite{ha:mi:09} using the generalized correlation for variable selection of linear models.

It is also worthy noticing that the threshold is a global measure (implicitly) depending on all $J$ variables. If we remove some $x$'s from the original explanatory variables set, the threshold value will be changed correspondingly. Thus the ``relevant'' and ``irrelevant'' regressors will also change.

\subsection{Cluster}
Motivated by the fact that in a block diagonal matrix, the nonzero entries along the diagonal are denser than those in the off-diagonal region and the assumption w.r.t. equation \eqref{eq:semi}: ``$\forall j \neq l$, $x_j \in \mathcal A_j$, $x_l \in \mathcal A_l$, $x_j$ and $x_l$ are (conditionally) independent given other $x$'s'', we define the following ``averaged non-zero'' score $S_{\mathcal{A}}$ for a index set $\mathcal{A}$: $S_{\mathcal{A}}\defeq \sum_{i,j \in \mathcal A}{\IF(\tilde \sigma_{ij}\neq0)}/|\mathcal A|^2$. Here we do not distinguish between the positive and negative values of $\tilde \sigma_{ij}$ since they could also be reflected by the corresponding linear coefficients as in equation \eqref{eq:semi}.
\begin{itemize}
\item[2] Perform the label permutation procedure for $x_1, \ldots, x_{K}$ to form clusters of (explanatory) variables (or $\mathcal A_1, \ldots, \mathcal A_S$) by utilizing the {``averaged non-zero''} score $S_\mathcal{A}$.
  \begin{itemize}
\item[2.1] Rank (in decreasing order) and relabel all $x_1, \ldots,x_k, \ldots, x_K$ according to $\sum_{1\leqslant j \leqslant {K}} {\IF(\tilde \sigma_{kj}\neq0)}$ to obtain the ``new'' $x_1, \ldots, x_K$. Always assume $x_1$ is in the first block (index set) $\mathcal A_1$;
\item[2.2] \textit{Forward} Include $x_k$ ($2 \leqslant k \leqslant K$) in the first index set ($x_k \in \mathcal A_1$) if $S_{{\mathcal A_1}\bigcup \{x_j\}} \geqslant S_{\mathcal A_1}$, and continue searching until the $K$th variable $x_K$. Without loss of generality (otherwise just relabel them), we assume $x_1, x_2, \ldots, x_{k-1} \in \mathcal A_1$.
   \item[2.3a] (For the case of no overlapping indices among $\mathcal A_1, \ldots, \mathcal A_S$) \newline        Given $\mathcal A_1$ formed in the last step, perform Steps 2.1 and 2.2 again for the variables not in the set $\mathcal A_1$, i.e. $\{1, 2, \ldots, K\} \backslash {\mathcal A_1}$ and construct $\mathcal A_2$.
          \item[2.3b] \textit{Backward} (replace Step 2.3a, for the case allowing overlapping indices among $\mathcal A_1, \ldots, \mathcal A_S$) \newline Given $\mathcal A_1$, perform Step 2.1 again for the variables not in the set $\mathcal A_1$, i.e. $\{1, 2, \ldots, K\} \backslash {\mathcal A_1}$ and start to construct $\mathcal A_2$, for example, $x_k \in \mathcal A_2$. Let $x_1 \in \mathcal A_1 \bigcap \mathcal \mathcal A_2$ only if $S_{\{x_1\} \bigcup \mathcal A_2} \geqslant S_{\mathcal A_2}$ and continue searching until
$x_{k-1}$. Notice that it is impossible for all $x_1, \ldots, x_{k-1} \in \mathcal A_2$ because of the way we construct $\mathcal A_1$ in the \textit{forward} step. Continue to construct $\mathcal A_2$ as in the \textit{forward} step by selecting variables w.r.t. $\{1, 2, \ldots, K\} \backslash {\mathcal A_1} \bigcup \{x_k\}$.
 \item[2.4] Continue this procedure until all variables $x_1, \ldots, x_K$ have been included into some index set(s) of $\mathcal A_1, \ldots, \mathcal A_S$, where $S$ is the number of selected index sets and $\mathcal A_1\bigcup \ldots \bigcup \mathcal A_S = \{1, 2, \ldots, K\}$. Given these, construct the corresponding semiparametric models by equation \eqref{eq:semi}.
    \end{itemize}
\end{itemize}

At Step $2$, if we can permute the variables' labels to have a block diagonal structure for the partition of the consistently estimated covariance matrix, such as the one in Figure \ref{multipleindex} (right), we can construct the corresponding class of semiparametric models as specific cases of equation \eqref{eq:semi}. By this step, we are grouping the (explanatory) variables into highly correlated groups, which are, however, weakly correlated with each other. This ``independence'' property (between the ``new'' predictors $\beta_s^\T x_{\mathcal A_s}, 1 \leqslant s \leqslant S$) is actually also required for the groupwise dimension reduction method of \cite{li:li:zh:10} for estimation. Except that we use the Spearman's rank correlation instead of the Pearson's correlation for ``screening'', a second difference between this work and \cite{fa:lv:08} is that we consider the covariance (correlation) matrix for all $x_1, \ldots, x_{J-1}, y$ variables instead of just between $y$ and $x_1, \ldots, x_{J-1}$, which is because we need to further group the relevant explanatory variables for semiparametric model construction at the second step.

A very important feature of the proposed label permutation procedure is that it is based on the thresholding regularized covariance matrix instead of the sample one. A related work which tries to discover the ordering of the variables through the metric multi-dimensional scaling method could be found in \cite{wa:le:09} for the i.i.d. Gaussian case. Their ultimate goal is to improve covariance matrix estimation rather than order the variables itself. Thus by utilizing the discovered ``order'' based on the sample covariance matrix, they estimate the large covariance matrix through banding regularization to enjoy the benefits brought by ordering. But in the case of large panels of economic and financial variables as we consider here, our ultimate goal is to cluster the variables to construct the proper semiparametric models instead of ``ordering''. For example, in the multiple index model, the order of the first index (or first cluster of variables) and the second index (or second cluster of variables) and the order of variables inside each ``cluster'' are both unimportant.

This case is also related to the hierarchical clustering, k-means algorithm and correlation clustering problem in computer science (\cite{de:im:03}, \cite{ba:bl:ch:04}), which aims to
partition a weighted graph with positive and negative edge weights so that negative edges
are broken up and positive edges are kept together. However, the correlation clustering algorithm
is also based on the sample correlation(s), and has also been shown to be NP-hard. Thus, as a \textit{key} difference with other works in the literature, instead of using the sample covariance (or correlation) matrix for ordering and clustering as \cite{wa:le:09}, \cite{de:im:03} and \cite{ba:bl:ch:04} did, we implement thresholding regularization for the sample covariance matrix and screening first and then find the corresponding groups through the stepwise label permutation procedure. It is simpler to be implemented than their's, since the thresholding regularized covariance matrix only has limited number of nonzero entries. By doing so, w.r.t. the regression setup, we also simultaneously extract the ``relevant'' explanatory variables for $y$ (Step $1$). Thus, we actually combine \textit{dimension reduction} and \textit{variable clustering}, which is especially suitable for modeling high dimensional data via semiparametric methods.

This procedure is computationally simple for a typical $J \leqslant 150$ macroeconomic and financial data set since the thresholding regularization procedure removes $J-1-K$ ``irrelevant'' variables first, and then rank the remaining $K$ ones before entering this label permutation procedure. Thus we avoid the NP-hard correlation clustering problem based on the sample covariance matrix.

\subsection{Estimate}
\begin{itemize}
\item[3] Groupwise dimension reduction with sign constraints.
\end{itemize}
For Step $3$, we implement the groupwise dimension reduction estimation procedure modified from \cite{li:li:zh:10}. If we implement their method directly, as we can see from Table \ref{tab:cpi} (details of data presented later), the Spearman's rank correlations between $x_1, \ldots, x_K$ and $y$ are all positive, however, some of their corresponding parametric coefficients are estimated to be \textit{negative} (details presented later in Table \ref{tab:a1g1}). This means that the consumer price index negatively depends on them, which is unlikely to be true from an economic point of view. Since we have disjoint groups of variables here and given the meaning of the Spearman's rank correlation, ideally, the sign of the corresponding parametric coefficient estimate w.r.t. $x_k, 1\leqslant k \leqslant K$ should be the same as the sign of the corresponding Spearman's rank correlation. This motivates us to add the \textit{sign constraint}, as a refinement, to the groupwise dimension reduction method developed in \cite{li:li:zh:10} to secure the \textit{sign consistency}.

Let us first consider a simple linear regression model $\E y = x_1 \beta_1 + , \ldots, + x_K\beta_K \defeq x^\T \beta$ with the constraint $\beta_1, \ldots, \beta_K \geqslant 0$. The linear coefficient $\beta$ could be estimated as the minimizer of $\|y-x^\T \beta\|_2^2/2 - \sum_{k=1}^K \lambda_k \beta_k$ with the corresponding nonnegative Lagrange multipliers $\lambda_k$'s, $1 \leqslant k \leqslant K$. If we denote $\diag[\lambda_1, \ldots, \lambda_K]$ by $\Lambda$, $\hat \beta = (x^\T x)^{-1}(x^\T y + \Lambda) \defeq \hat \beta_{OLS} + (x^\T x)^{-1} \Lambda$. Intuitively, in case some entry of $\hat \beta_{OLS}$, say the $k$th, is negative, which contradicts the initial requirement $\beta_k \geqslant 0$, $(x^\T x)^{-1} \lambda_k$ plays the role of adding a positive increment to it, s.t. $\hat \beta_k \geqslant 0$.

Similarly, in our setup, if we use $\tilde \sigma_{kJ}$ to denote the Spearman's rank correlation estimate between $x_{k}$ and y ($x_J$) extracted from $T_{\hat s}(\hat \Sigma)$ and add the sign constraint $sign(\tilde \sigma_{kJ})\beta_{k} \geqslant 0$ to the estimation procedure of \cite{li:li:zh:10} ($\beta_s$ here corresponds to their $\beta_g$), a simple calculation shows that we just need replace their estimation equation (15) for $\beta \defeq (\beta_1, \ldots, \beta_K)^\T$ by ($\beta$ here corresponds to their $\zeta$):
\begin{align}
\hat \beta &= \Big\{\sum_{i=1}^T \sum_{j=1}^T {\bf R}^{ij} {\bf R}^{ij\T} K_h(V^j - V^i)\Big\}^{-1} \Big\{\sum_{i=1}^n \sum_{j=1}^n (Y^j -a^i) K_h(V^j - V^i) {\bf R}^{ij} + \Lambda' \Big\},\nonumber \\
&= \hat \zeta +\Big\{\sum_{i=1}^T \sum_{j=1}^T {\bf R}^{ij} {\bf R}^{ij\T} K_h(V^j - V^i)\Big\}^{-1}\Lambda' \label{zetaest}
\end{align}
where $\Lambda'$ is a $K \times K$ diagonal matrix $\diag[\lambda_1 sign(\tilde \sigma_{1J}), \ldots,  \lambda_K sign(\tilde \sigma_{KJ})]$; $\{\lambda_{k}, 1 \leqslant k \leqslant K\}$ are the hyperparameters; $\hat \zeta$ and other variables are the same as in equation (15) of \cite{li:li:zh:10}. In general, selection of $\lambda$'s requires minimizing some loss function. Motivated by the discussion above for the simple linear regression case, when $sign(\hat \zeta_k)$ is the same as $sign(\tilde \sigma_{kJ})$, we simply choose $\lambda_{k}=0$, otherwise choose $\lambda_{k}$ to be the minimum (positive) value s.t. $\hat \beta_k=0$. By our experience, this works well and the convergence of the iterative estimation procedure is achieved within $19$ iteration steps ($10^{-6}$ as the tolerance) for modeling CPI, which is to be presented in Section \ref{application}. Then by the property of the convex minimization problem, if a local minimum exists, it is also a global minimum.

Overall, similar to \cite{fa:lv:08}'s ``screen first; fit later'' approach for modeling high dimensional data, ours could be considered as the ``screen first; group second; fit third'' approach. Alternatively, \cite{2008arXiv0801.1095B} and \cite{me:bu:06} consider the ``fit first; screen later'' approach. In general, a great deal of work is needed to compare ``screen first; fit later'' type of methods with ``fit first; screen later'' types of method in terms of consistency and oracle properties. But when the spatial structure is complex (thus we need deviate from linearity), in terms of semiparametric modeling, as we have discussed in Section \ref{intro}, the later one might face several main limitations, while ours, together with the estimation method modified from \cite{li:li:zh:10}, as a special case of the former one, could circumvent these issues and would be faster when dealing with higher dimensionality.

\section{Application}\label{application}
We use the dataset of \cite{sto:wat:05a}. This dataset contains $131$ monthly macro indicators covering a broad range of categories including income, industrial production, capacity, employment and unemployment, consumer prices, producer prices, wages, housing starts, inventories and orders, stock prices, interest rates for different maturities, exchange rates, money aggregates and so on. The time span is from January $1959$ to December $2003$. We apply logarithms to most of the series except those already expressed in rates. The series are transformed to obtain stationarity by taking (the $1$st or $2$nd order) differences of the raw data series (or the logarithm of the raw series). Then all observations are standardized.

\begin{figure}
\centering
  \includegraphics[width=7.5cm]{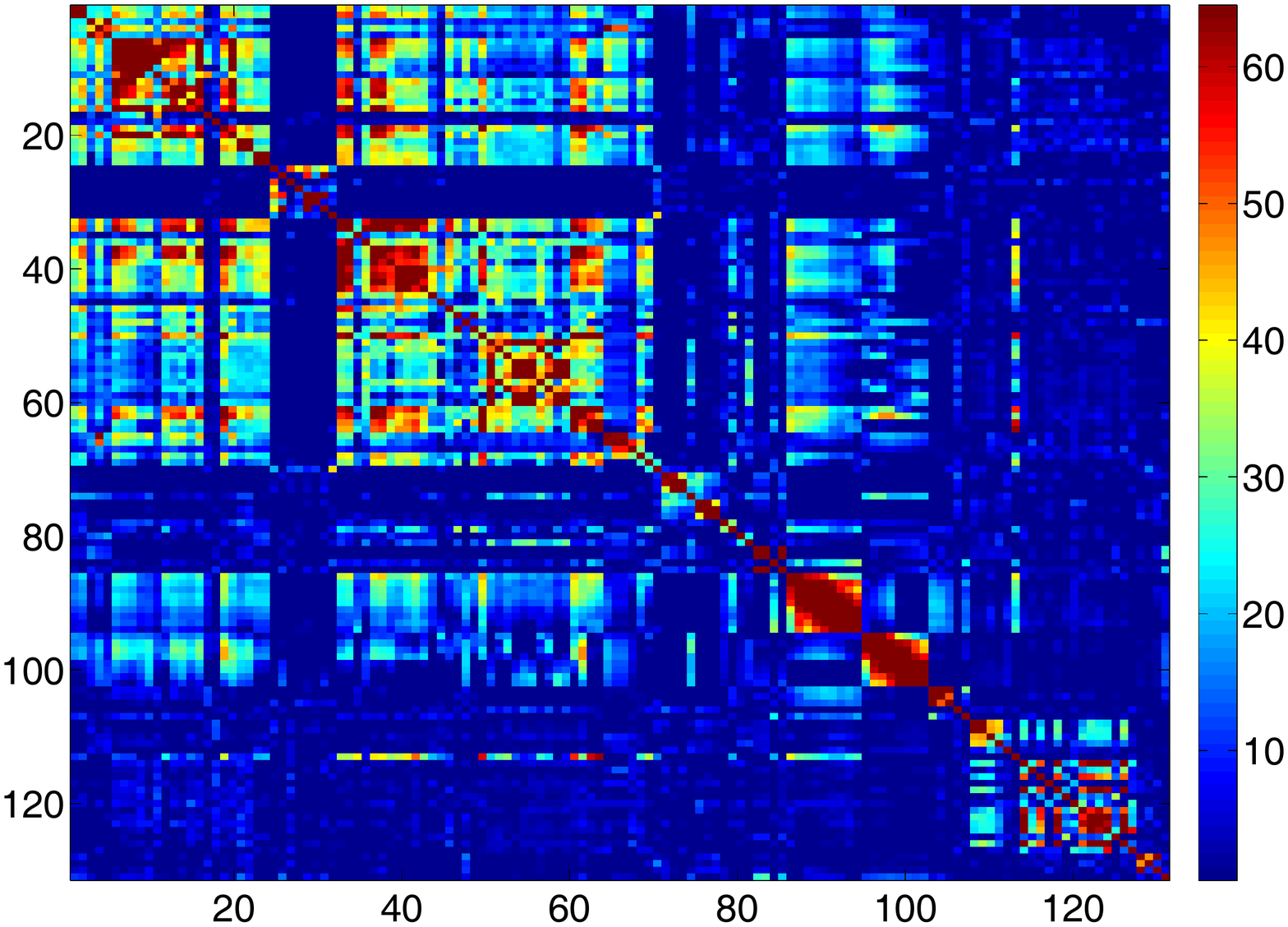} \includegraphics[width=7.5cm]{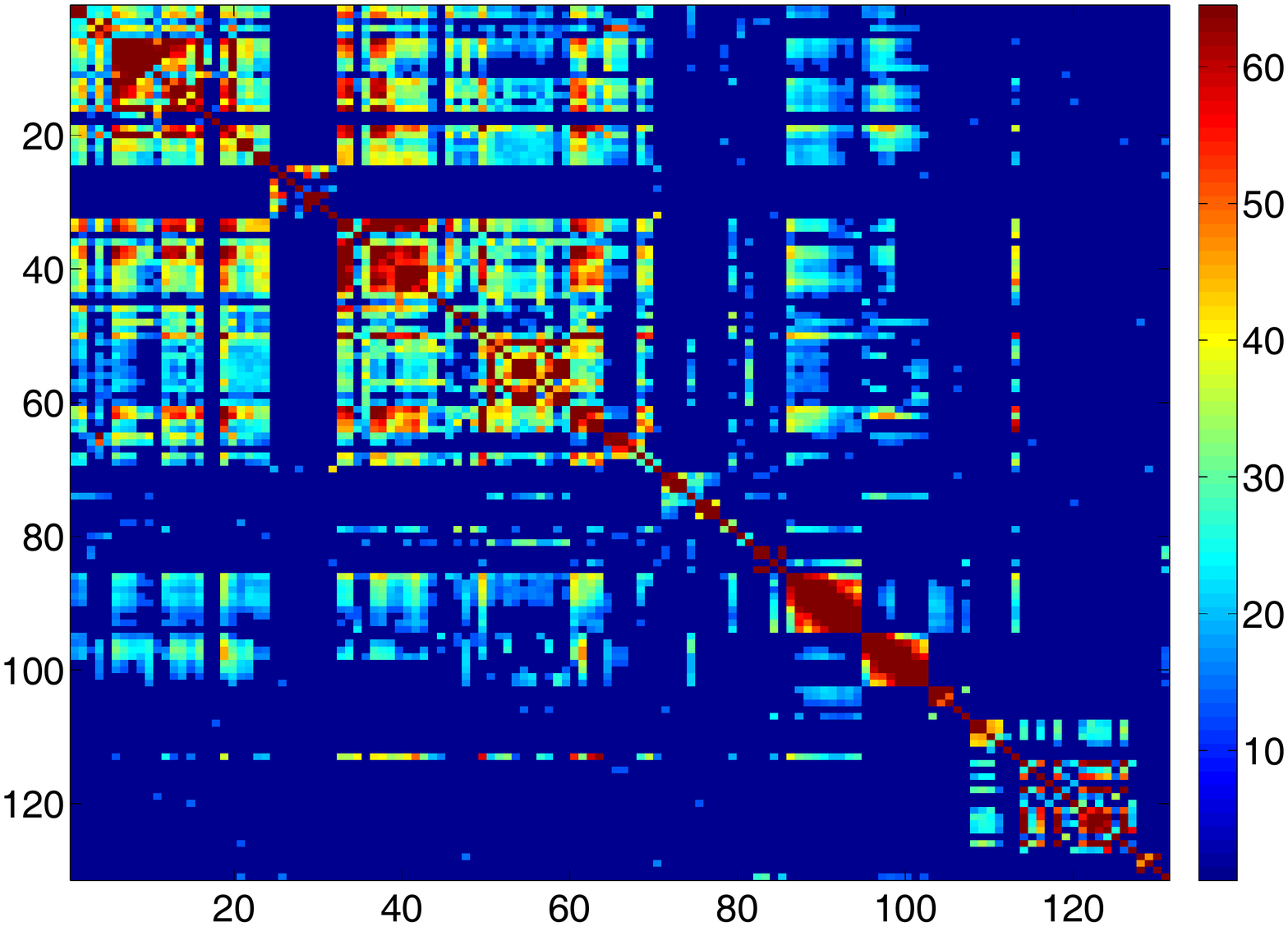}
  \caption{Sample and regularized covariance matrices (after multiplying each entry's value by $100$).}\label{samcov}
\end{figure}

\begin{table}[h]
\centering
{\scriptsize\begin{tabular}{ccccccccccccccccc}
\hline \hline
\bf CPI	&	0.62	&	0.76	&	0.65	&	0.65	&	0.51	&	0.67	&	0.65	&	0.25	&	0.28	&	 0.28	&	0.27	&	0.17	&	0.25	&	0.14	&	0.25	\\ \hline
	&	\bf 121	&	0.61	&	0.55	&	0.51	&	0.56	&	0.52	&	0.51	&	0.17	&	0.22	&	 0.27	&	0.19	&	0.20	&	0.22	&	0.14	&	0.32	\\
	&		&	\bf 123	&	0.66	&	0.62	&	0.47	&	0.73	&	0.65	&	0.24	&	0.26	&	0.26	 &	0.25	&	0.19	&	0.22	&	0.13	&	0.25	\\
	&		&		&	\bf 122	&	0.65	&	0.55	&	0.70	&	0.71	&	0.29	&	0.28	&	0.24	&	 0.23	&	0.13	&	0.21	&	0	&	0	\\
	&		&		&		&	\bf 124	&	0.45	&	0.63	&	0.80	&	0.23	&	0.25	&	0.26	&	 0.30	&	0	&	0.58	&	0.26	&	0	\\
	&		&		&		&		&	\bf 116	&	0.54	&	0.51	&	0.20	&	0.25	&	0.24	&	0	&	 0.19	&	0	&	0.15	&	0	\\
	&		&		&		&		&		&	\bf 118	&	0.77	&	0.29	&	0.31	&	0.29	&	0.27	&	 0.19	&	0	&	0	&	0	\\
	&		&		&		&		&		&		&	\bf 126	&	0.31	&	0.35	&	0.30	&	0.34	&	0	 &	0.14	&	0	&	0	\\
	&		&		&		&		&		&		&		&	\bf 108	&	0.87	&	0.46	&	0	&	0	&	0	 &	0	&	0	\\
	&		&		&		&		&		&		&		&		&	\bf 109	&	0.47	&	0	&	0	&	0	&	 0	&	0	\\
	&		&		&		&		&		&		&		&		&		&	\bf 110	&	0	&	0	&	0	&	0	 &	0	\\
	&		&		&		&		&		&		&		&		&		&		&	\bf 115	&	0	&	0	&	0	 &	0	\\
	&		&		&		&		&		&		&		&		&		&		&		&	\bf 119	&	0	&	 0.35	&	0	\\
	&		&		&		&		&		&		&		&		&		&		&		&		&	\bf 127	&	0	 &	0.27	\\
	&		&		&		&		&		&		&		&		&		&		&		&		&		&	\bf 125	&	0	\\
	&		&		&		&		&		&		&		&		&		&		&		&		&		&		&	 \bf 120	\\
\hline \hline
\end{tabular}}
\caption{Partition of the $T_s(\hat \Sigma_{1,v})$ w.r.t. $CPI$ and the $15$ ``relevant'' variables and the diagonal entries denote the indices (in the original data set) of the corresponding variables.} \label{tab:cpi}
\end{table}
\begin{table}[h]
\centering \begin{tabular}{ll r r}
   \hline   \hline
   Variable & Meaning& Coefficients  & Coefficients\\
   \hline
  $PWFCSA_{108}$ & Producer Price Index: Finished Goods &$0.150$& $0.207$\\
  $PWIMSA_{109}$ & PPI: Finished Consumer Goods &$0.741$& $0.763$\\
  $PWCMSA_{110}$ & PPI: Intermed. Mat. Supplies \& Components &$0.074$& $0.196$\\
  $PU84_{116}$ & CPI-U: Transportation &$0.307$& $0.304$\\
  $PUC_{118}$ & CPI-U: Commodities &$0.290$& $0.235$\\
  $PUXF_{121}$ & CPI-U: All Items Less Food &$0.397$& $0.378$\\
  $PUXHS_{122}$ & CPI-U: All Items Less Shelter &$-0.034$& $0$\\
  $PUXM_{123}$ & CPI-U: All Items Less Medical Care &$0.246$& $0.217$\\
  $GMDC_{124}$ & PCE,IMPL PR DEFL:PCE &$-0.146$& $0$\\
  $GMDCN_{126}$ & PCE,IMPL PR DEFL:PCE; Nondurables &$-0.061$& $0$\\
\hline
$PUCD_{119}$ & CPI-U: Durables &$-0.639$&$0.635$\\
$GMDCD_{125}$& PCE,IMPL PR DEFL:PCE; Durables &$-0.770$&$0.772$\\
\hline
$PUS_{120}$& CPI-U: Services &$-0.994$& $0.979$\\
$GMDCS_{127}$& PCE,IMPL PR DEFL:PCE; Services &$0.111$& $0.203$\\
\hline
$PU83_{115}$& CPI-U: Apparel \& Upkeep &$1$ & $1$\\
  \hline   \hline
\end{tabular}
\caption{Detailed meanings of the variables with the corresponding parametric coefficients' estimates using the groupwise dimension reduction method without ($3$rd column) and with ($4$th column) sign constraints.}\label{tab:a1g1}
\end{table} 
Figure \ref{samcov} contains plots of the sample and thresholding regularized Spearman's rank correlation matrices based on the ``optimal'' threshold $0.13$ selected by the cross validation procedure discussed in subsection \ref{sub:threscv} with $T_1=120, T_2=240$. The variables of special interest include the consumer price index (CPI) as a measure of prices and an economic indicator. The annual percentage change in CPI is used as a measure of inflation. CPI can be used to index (i.e., adjust for the effect of inflation) the real value of wages, salaries and pensions, and also for regulating prices and deflating monetary magnitudes to show changes in real values. Besides being a deflator of other economic series, it is also a means of adjusting dollar values. Thus CPI is one of the most closely watched national economic statistics. To this end, we use modeling of CPI to illustrate our method. Table \ref{tab:cpi} displays the partition of the $T_s(\hat \Sigma_{1,v})$ w.r.t. the variables relevant to CPI. The detailed meanings of these variables (and corresponding three digits indices in the data set provided by \cite{sto:wat:05a}) are given in Table \ref{tab:a1g1}. By the forward (and backward) procedure discussed in Section \ref{permutation}, we find the following index sets for constructing semiparametric models for modeling CPI. The one without overlapping is presented on the LHS, and that allowing overlapping is presented on the RHS.
\begin{align*}
\mathcal A_1 &=  108 - 110,  116,   118, 121 - 124,     126 &\mathcal A'_1 &=  108 - 110,  116,   118, 121 - 124,     126\\
\mathcal A_2 &=  119, 125 & \mathcal A'_2 &=  121 - 124,   115\\
\mathcal A_3 &=  120, 127 & \mathcal A'_3 &=  121 - 123,   119\\
\mathcal A_4 &=115 & \mathcal A'_4 &=121 - 123,   127\\
 && \mathcal A'_5 &=121, 123, 125\\
&& \mathcal A'_6 &=120, 121, 123
\end{align*}
Notice that $\mathcal A'_1=\mathcal A_1$, and the main difference between these two methods comes from $\mathcal A_2 - \mathcal A_4$ and $\mathcal A'_2 - \mathcal A'_6$, i.e. how to allocate the $119, 125, 120, 127, 115$th variables, which originally results from the rank correlations between them and the $121-124$th variables. As we can see from Table \ref{tab:a1g1}, the $121-124$th variables are very close to the $y$ variable: CPI-U: All Items (82-84=100, SA) except one item (food, shelter or medical care) or the implicit price deflator (of personal consumption expenditures).

Due to the identification and estimation problems we discussed before, from now on, we mainly concentrate on the disjoint index sets case and suggest the following semiparametric model for modeling CPI: $\E(CPI_{114})=$
{\small\begin{align}
&\nonumber \\&g_1\Big(\beta_{108} PWFCSA_{108} + \beta_{109} PWIMSA_{109} +\beta_{110} PWCMSA_{110}+  \beta_{116} PU84_{116}+\beta_{118} PUC_{118}\nonumber \\&+\beta_{121}PUXF_{121}+\beta_{122} PUXHS_{122} +\beta_{123}PUXM_{123}+\beta_{124} GMDC_{124}+\beta_{126} GMDCN_{126}\Big)\nonumber\\& + g_2\Big(\beta_{119}PUCD_{119}+\beta_{125} GMDCD_{125}\Big) +g_3\Big(\beta_{120} PUS_{120} + \beta_{127} GMDCS_{127}\Big)+g_4\Big(PU83_{115}\Big),\label{eq:cpi}
\end{align}}
where $g_1, \ldots, g_4$ are the unknown link functions to be estimated nonparametrically and $\beta_{108}, \ldots, \beta_{127}$ are unknown parameters which belong to the parameter space.
The variables $PUCD_{119}$ and $GMDCD_{125}$ denote the consumer price index and implicit price deflator (of personal consumption expenditures) for durable goods respectively. Thus $\mathcal A_2$ could be interpreted as the index set for durable goods. Very similarly, $\mathcal A_3$ and $\mathcal A_4$ could be interpreted as the index sets for service, and apparel and upkeep respectively which are also important factors affecting consumer price index. All common factors strongly associated with CPI are included in $\mathcal A_1$. Compared with the linear, additive or single index models, the model \eqref{eq:cpi} actually combines flexibility in statistical modeling and interpretability from an economic point of view, while being kept close to the data's complex spatial structure.

We further employ the groupwise dimension reduction method in \cite{li:li:zh:10} to estimate \eqref{eq:cpi} and present the parametric coefficients' estimate in Table \ref{tab:cpi} ($3$rd column). Contradicting to the background knowledge of economics and the positive Spearman's rank correlations between CPI and $PUXHS_{122}$, $GMDC_{124}$, $GMDCN_{126}$, $PUCD_{119}$, $GMDCD_{125}$, $PUS_{120}$ shown in Table \ref{tab:a1g1}, their corresponding parametric coefficients are estimated to be \textit{negative}. This means that the consumer price index negatively depends on them, which is unlikely to be true. Finally we apply the modified procedure with sign constraints to estimate \eqref{eq:cpi} again and present the corresponding parametric coefficients' estimates in the last column of Table \ref{tab:a1g1} with the explained variation $85.8\%$. While $\beta_{119}$, $\beta_{125}$ and $\beta_{120}$ are estimated positively, $\beta_{122}$, $\beta_{124}$ and $\beta_{126}$ are estimated to be $0$, which means $PUXHS_{122}$, $GMDC_{124}$ and $GMDCN_{126}$ could be eliminated from the model. To compare with the linear models and see the advantages of semiparametrics, we also consider the linear model using all other $130$ variables (except CPI itself). The explained variation w.r.t. the LARS estimate (least angle regression, developed by \cite{citeulike:3284841}) is $80.4\%$.

Besides the measure of prices, other variables of special interest include a measure of real economic activity  and a monetary policy instrument. As in \cite{Ch:Ei:Ev:99}, we use employment as an indicator of real economic activity measured by the number of employees on non-farm payrolls (EMPL). The monetary policy instrument is the Federal Funds Rate (FFR). If we apply the SCE approach to estimate EMPL and FFR, the explained variation is $99.9\%$ and $97.6\%$ respectively, while the corresponding LARS estimates' is $99.6\%$ and $86.7\%$. These results are summarized in Table \ref{seclars}. Thus we see that we could reduce the SSE approximately by $27.4\%$ for CPI and $82.0\%$ for FFR through considering the (flexible and proper) semiparametrics. The improvement for EMPL is not significant since the LARS estimate has already performed quite well.
\begin{table}[h]
  \centering
  \begin{tabular}{lccc}
 \hline \hline
 & CPI & EMPL & FFR \\
\hline
  $R^2$ (SCE) & $85.8\%$ & $99.9\%$ & $97.6\%$ \\
  $R^2$ (LARS) & $80.4\%$ & $99.6\%$ & $86.7\%$ \\
  \hline \hline
\end{tabular}
  \caption{Explained variation of the SCE and LARS estimates for CPI, EMPL and FFR.}\label{seclars}
\end{table}

\section{Concluding Remarks and Discussions}\label{discussion}
In this paper, we consider estimating a large spatial covariance matrix of the generalized $m$ dependent and $\beta$-mixing time series (with $J$ variables and $T$ observations) by hard thresholding regularization. We quantify the interplay between the estimators' consistency rate and the time dependence level, discuss an intuitive resampling scheme for threshold selection, and prove a general cross-validation result that justifies this approach. Given a consistently estimated large sparse covariance matrix, by utilizing the natural links
 among graphical models, semiparametrics and large spatial covariance matrix, we propose a novel forward (and backward) label permutation procedure to form a block diagonal structure for it and construct the corresponding low dimensional semiparametric model. Finally we apply this method to study the spatial structure of large panels of economic and financial time series to find the proper semiparametric structure for estimating the consumer price index (CPI) and present its superiority over the linear models.

\textit{Choice of Threshold}

Concerning the choice of threshold in the context of time series analysis, if we are mainly targeting estimation performance of the corresponding semiparametric models instead of minimizing the loss functions \eqref{emploss} and \eqref{oraloss} related to the covariance matrix estimation, we might directly consider minimizing the estimation error based on the selected semiparametric model, for example \eqref{eq:multipleindex}, s.t. the prediction performance might be optimized.

\textit{Other Measures of Dependence for Screening}

The information given by a Pearson's correlation coefficient is not enough to define the dependence structure between random
variables. Except the Spearman's rank correlation we used here, distance correlation, \cite{sz:ri:ba:07} and Brownian covariance (correlation), \cite{sz:ri:09} were also introduced to address the deficiency of Pearson's correlation that it can be zero for dependent random variables; zero distance correlation and zero Brownian correlation imply independence. The correlation ratio is able to detect almost any functional dependency, and the entropy-based mutual information/total correlation is capable of detecting even more general dependencies. We want to point out that the Step $1$ of the SCE procedure could be very easily extended to these measures above and the threshold value could be selected by the cross-validation procedure similarly.

It is also noteworthy that \cite{fa:fe:so:11} considers the independence screening procedure by ranking the explanatory variable's importance according to the descent order of the residual sum of squares of the componentwise
nonparametric regressions or the marginal strength of the marginal nonparametric regression. By doing that, they (implicitly) assume that the true semiparametric structure is additive, which is different from our ultimate goal here: construct the proper semiparametric structure.

\textit{Theoretical Study of the Screening Step}

Noticing that $F_i(x_i)$ and $F_j(x_j)$ in the formula \eqref{def:rank} follow the uniform distribution on $[0, 1]$, thus \eqref{def:rank} could be simplified as $\rho_{x_i, x_j}=12 \E\{F_i(x_i)F_j(x_j)\}-3$. Similar to the ``sure independence screening'' property of \cite{fa:lv:08} using Pearson's correlation for variable screening of linear models, to study the theoretical property of the screening step here based on the Spearman's rank correlation, parallel to the equation (20) ``$\omega=X^\T y = X^\T X \beta+ X^\T \eps$'' of \cite{fa:lv:08}, we could define
\begin{align}
\omega &=(\omega_1, \ldots, \omega_{J-1})\nonumber\\
\omega_j &= 12 F_j(x_j)F_J(x_J)-3 \defeq 12 F_j(x_j)F_y(y) -3, \quad 1 \leqslant j \leqslant J-1 \nonumber \\
&=12 F_j(x_j)F_y\Big\{\sum_{s=1}^S g_s(\beta_s^\T x_{\mathcal A_s}) + \eps_0\Big\} -3,  \label{eq:omega}
\end{align}
where $\eps_0$ is the (conditional) mean-zero error term from approximating $y$ by $\sum_{s=1}^S g_s(  \beta_s^\T x_{\mathcal A_s})$ in \eqref{eq:semi}. Similar to the idea of the (group) MAVE method of \cite{xi:to:li:zh:02}, \cite{li:li:zh:10}, we notice that
\begin{equation*}
\partial g_s (\beta_s^\T x_{\mathcal A_s})/\partial x_{\mathcal A_s} = g'_s(\beta_s^\T x_{\mathcal A_s}) \beta_s,
\end{equation*}
provided by $g'_s(\beta_s^\T x_{\mathcal A_s})$ is well defined. Thus applying the Taylor expansion to $\sum_{s=1}^S g_s(\beta_s^\T x_{\mathcal A_s}) + \eps_0$ at $x'$ will help linearize it as:
\begin{align*}
&a + \sum_{s=1}^S g'_s(\beta_s^\T x'_{\mathcal A_s}) \beta_s^\T (x-x')_{\mathcal A_s} + \CO \{\sum_{s=1}^S(x-x')_{\mathcal A_s}^\T(x-x')_{\mathcal A_s}\} + \eps_0 \\
\defeq & a + \sum_{s=1}^S b_s \beta_s^\T (x-x')_{\mathcal A_s} +\eps.
\end{align*}
Therefore, we could rewrite \eqref{eq:omega} as
\begin{equation}
12 F_j(x_j)F_y\Big\{a + \sum_{s=1}^S b_s \beta_s^\T (x-x')_{\mathcal A_s} +\eps\Big\} -3. \label{eq:omega2}
\end{equation}
Studying the property of \eqref{eq:omega2} will be the main focus. However, due to the presence of the cumulative density functions $F_j$ and $F_y$ here, this is expected to be much more complex than the Pearson's correlation case. We hope that the other people could further investigate this.

\vspace{1cm}
{\textbf{Acknowledgement}} This work was partially motivated during a conversation with Prof Lixing Zhu in Hong Kong in Feb, $2010$. The author is very grateful to Prof Peter Bickel, Prof Lixing Zhu, Prof Lexin Li, Mu Cai and Ying Xu for very interesting discussions and comments on this and related topics. In particular, I would like to thank Prof Peter Bickel for sponsoring my stay at the University of California, Berkeley.

\section{Appendix}\label{appendix}
\noindent\textbf{Proof of Theorem \ref{dependent}}
The proof of this theorem is based on the ones of Theorem 1 and 2 in \cite{Bickel08regularizedestimation2} up to a modification of the bound on $\P\{\max_{i,j} |\hat \sigma_{ij}-\sigma_{ij}|\geqslant s\}$, as remarked by their subsection 2.3. By the definition of $\hat \sigma_{ij}$ in \eqref{def:sample} and the assumption that for all $i$ and $j$, $|X_{ti}X_{tj}| \leqslant M_t$ holds with a high probability, applying the (extended) Mcdiarmid inequality, see Theorem $2.1$ of \cite{ja:04}, to the sum of dependent random vectors $\sum_{t=1}^T |X_{ti}X_{tj}|$ yields:
\begin{equation*}
 \P\{\max_{i,j} |\hat \sigma_{ij}-\sigma_{ij}|\geqslant s\}\leqslant J^2 \exp\left\{-\frac{s^2 T^2}{\cx^*(\ct) \sum_t M_t^2}\right\}=\exp\{(2-M'^2)\log J\},
\end{equation*}
where $s_T= M' \sqrt{{{\log J \, \cx^*(\ct)}}/{T}}$ with sufficiently large $M'$ also depending on $C'$ with $\sum_{t=1}^T M_t^2 C'/T \leqslant C'$ and ${{\log J \, \cx^*(\ct)}}/{T} = \Co(1)$. Since equation (10) in \cite{Bickel08regularizedestimation2} holds, others go through verbatim. This completes the proof. \hfill $\qquad\square$

\addvspace{3ex}
\noindent\textbf{Proof of Theorem \ref{theo:mixing}}
The proof of this theorem is also based on the ones of Theorem 1 and 2 in \cite{Bickel08regularizedestimation2} up to a modification of the bound on $\P\{\max_{i,j} |\hat \sigma_{ij}-\sigma_{ij}|\geqslant s\}$. Assume the $\beta$-mixing sequence $\{X_{ti}X_{tj}\}_{t=1}^T$ to satisfy Assumption \ref{ass:mixing} $\forall i, j$, applying the Bernstein type inequality for $\beta$-mixing random variables $\{X_{ti}X_{tj}\}_{t=1}^T$, see Theorem $4$ of \cite{do:94}[P.36], yields that, $\forall \eps>0$ ($\theta \defeq \eps^2/4$) and $\forall$ $0 < q \leqslant 1$,
\begin{equation}
\hspace{-0.3cm}\P(|\sum_{t=1}^T X_{ti}X_{tj}|\geqslant sT) \leqslant \underbrace{4 \exp\Big[-\frac{(1-\eps)3(1+\theta)s^2 T}{2\{ 3(1+\theta)\sigma^2+qMsT\}}\Big]}_{\defeq A} +\underbrace{2\frac{(1+\theta) \beta_{mix}}{q}}_{\defeq B}.
\label{mixing:prob}
\end{equation}
To make $J^2 (A+B)$ arbitrarily small, we choose $s_T= M' \sqrt{\frac{\log J}{T}}$ with sufficiently large $M'$ also depending on $\eps, \sigma^2, M$, $\log J / T = \Co(1)$, $q=3(1+\theta)\sigma^2/(MsT)$, and {$\beta_{mix}=\CO\{(J^{2+\delta'} \sqrt{\log J T})^{-1}\}$} with $\delta' >0$. Thus $A$ and $B$ are bounded by $\exp(-M'^2 \log J)$ and $J^{-(2+\delta')}$ respectively, which can be arbitrarily close to $0$. This completes the proof. \hfill $\qquad\square$

\addvspace{3ex}
\noindent\textbf{Proof of Lemma \ref{covlemma3}}
Since
$$\P\Big(J^{-1}|tr(V {\hat \Sigma_B} - V\Sigma)|\geqslant s \Big) \leqslant \P\Big(J^{-1}|tr(B^{-1} {\sum_{p=1}^B V X_p X_p^\T} - V\Sigma)|\geqslant s \Big),$$
and $tr(X_p X_p^\T)=tr(X_p^\T X_p)$, $tr({\sum_{p=1}^B V X_p X_p^\T} - V\Sigma) \leqslant J \sum_{p=1}^B \bar X_{p}^2$ with $\bar X_{p}=J^{-1} \sum_{j=1}^J X_{pj}$, applying the same inequality as in \eqref{mixing:prob} to $\sum_{p=1}^B \bar X_{p}^2$ leads to $$\P\Big(J^{-1}|tr(V {\hat \Sigma_B} - V\Sigma)|\geqslant s \Big) \leqslant K_1 \exp(-K_2 s^2 B)$$
with some constants $K_1$ and $K_2$.

Consequently we also have
\begin{equation}
\P\Big(J^{-1}\max_{p=1, \ldots, P} |tr(V_p {\hat \Sigma_B} - V_p\Sigma)|\geqslant s \Big)  \leqslant \IF (0 \leqslant s \leqslant x) + K_1 P \exp(-K_2 s^2 B) \IF(s>x).\label{eq:57}
\end{equation}
If we integrate \eqref{eq:57}, i.e.
\begin{align}
J^{-1}\E \max_{p=1, \ldots, P} \Big(|tr\{v_j {\hat \Sigma_B} - \E (v_j\Sigma)\}| \Big) \leqslant x + K_1 P \int_x^\infty \exp(-K_2 s^2 B) ds,  \label{eq:58}
\end{align}
and minimize the RHS of \eqref{eq:58} over $x$ as $P \rightarrow \infty$, we find that the minimizer satisfies \newline $x=C(q, c_0, M)\sqrt{{\log P}/{B}} \{1+ \Co(1)\}$. Hence
$$J^{-1}\E \max_{p=1, \ldots, P} \Big(|tr\{v_j {\hat \Sigma_B} - \E (v_j\Sigma)\}| \Big) \leqslant C(q, c_0, M)\sqrt{{\log P}/{B}}. \hfill \qquad\square$$

\addvspace{3ex}
\noindent\textbf{Proof of Theorem \ref{th:cv}}
Based on Lemma \ref{covlemma3}, we conclude that $\rho(P)$ from the second condition of Lemma \ref{cvpreparation} satisfies
$$\rho(P) \leqslant C(q, c_0, M) J^2 {\log P}/{B}.$$
Hence, Lemma \ref{covlemma3} implies that
$$\E \|B^{-1} \sum_{p=1}^B X_p X_p^\T - \Sigma\|_v \leqslant C_1 \rho(P).$$
Hence, if we select $B_T = T \varepsilon(T, J)$ and $\log P=\Co\{T^{q/2}c_0(J)J^{-1}(\log J)^{1-q/2}{\eps(T, J)}\}$, the conditions of Lemma \ref{cvpreparation} are satisfied and Theorem \ref{th:cv} follows. \hfill $\qquad\square$

\bibliography{biball}
\bibliographystyle{apalike}
\end{document}